%% file: main.tex
\documentclass[11pt]{article}

\usepackage[final]{acl}

\usepackage{times}
\usepackage{latexsym}

\usepackage[T1]{fontenc}

\usepackage[utf8]{inputenc}

\usepackage{microtype}

\usepackage{inconsolata}

\usepackage{graphicx}

\usepackage{kotex}
\usepackage{amsmath}
\usepackage{multirow}
\usepackage{arydshln}
\usepackage[normalem]{ulem}
\useunder{\uline}{\ul}{}
\usepackage{tcolorbox}
\usepackage{caption}
\usepackage{subcaption}
\usepackage{booktabs}
\usepackage{colortbl}
\usepackage{setspace}
\usepackage{tabularx}
\usepackage{alltt}
\usepackage{listings}
\usepackage{amssymb}
\usepackage{pifont}   
\usepackage{xcolor}   
\usepackage[dvipsnames]{xcolor}
\usepackage[table]{xcolor}
\usepackage{algorithm}
\usepackage{algpseudocode}
\usepackage{subcaption}
\usepackage{adjustbox}
\tcbuselibrary{breakable}

\newtcbox{\boxedtag}{on line,
  colframe=Blue,    
  colback=Orange!58, 
  boxrule=0.3pt,     
  arc=0pt,           
  boxsep=0.1pt,        
  left=1pt,
  right=1pt,
  top=0.5pt,
  bottom=0.5pt
}

\newcommand{\bgminipage}[2]{%
  \colorbox{#1!10}{\parbox{\dimexpr\linewidth-2\fboxsep\relax}{#2}}%
}

\title{ReFEree: Reference-Free and Fine-Grained Method for Evaluating Factual Consistency in Real-World Code Summarization}

\author{
 \textbf{Suyoung Bae},
 \textbf{CheolWon Na},
 \textbf{Jaehoon Lee},
 \textbf{Yumin Lee},
 \textbf{YunSeok Choi}\thanks{Corresponding authors},
 \textbf{Jee-Hyong Lee}$^*$
\\
 College of Computing and Informatics \\ Sungkyunkwan University, South Korea
\\
  \{sybae01, ncw0034, hoon1223, totoandy, ys.choi, john\}@skku.edu
}

\begin{document}
\maketitle
\begin{abstract}
As Large Language Models (LLMs) have become capable of generating long and descriptive code summaries, accurate and reliable evaluation of factual consistency has become a critical challenge. However, previous evaluation methods are primarily designed for short summaries of isolated code snippets. Consequently, they struggle to provide fine-grained evaluation of multi-sentence functionalities and fail to accurately assess dependency context commonly found in real-world code summaries.
To address this, we propose \textbf{ReFEree}, a reference-free and fine-grained method for evaluating factual consistency in real-world code summaries. We define factual inconsistency criteria specific to code summaries and evaluate them at the segment level using these criteria along with dependency information. These segment-level results are then aggregated into a fine-grained score. We construct a code summarization benchmark with human-annotated factual consistency labels. The evaluation results demonstrate that ReFEree achieves the highest correlation with human judgment among 13 baselines, improving 15-18\% over the previous state-of-the-art. Our code and data are available at \url{https://github.com/bsy99615/ReFEree.git}.
\end{abstract}

\input{texs/1_introduction_v3}
\input{texs/2_related}

\input{texs/3_method}

\input{texs/4_experiment_setups}

\input{texs/5_result_analysis_v2}

\input{texs/6_conclusion}

\bibliography{custom_ r}

\input{texs/7_appendix_v2}

\end{document}

%% file: texs/1_introduction_v3.tex
\section{Introduction}

\begin{figure}[t]
    \centering
    \includegraphics[width=\linewidth]{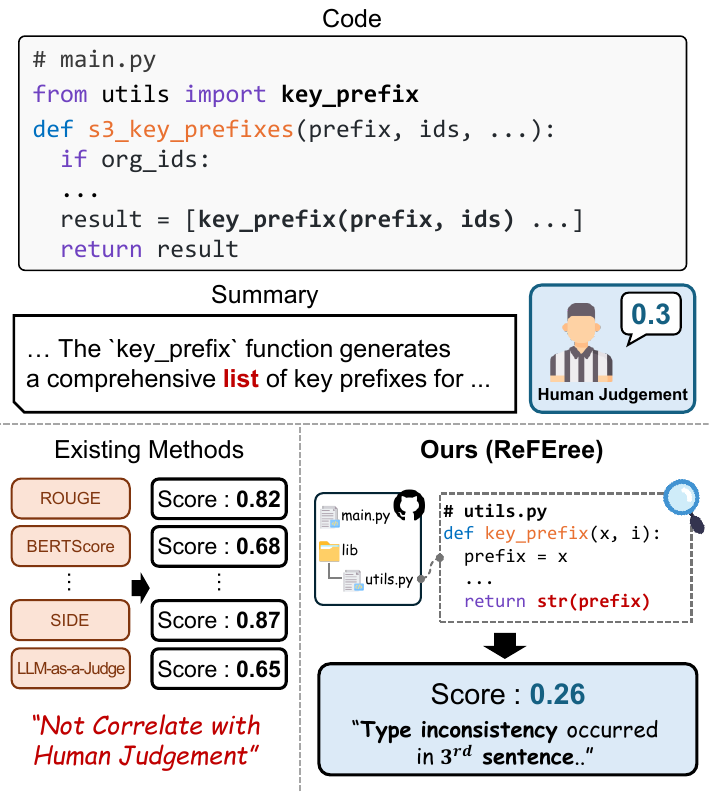}
    \caption{When evaluating factual consistency in real-world code summaries, existing methods (\textbf{bottom left}) fail to capture that the explanation of the \texttt{`key\_prefix'} in the summary exhibits a type error. In contrast, our method (\textbf{bottom right}) uses segment-level evaluation with related \texttt{`key\_prefix'} information, better aligning with human judgment.}
    \label{figure:intro}
\end{figure}

Recent advances in Large Language Models (LLMs), such as GPT-4, have made it feasible to automatically generate long and descriptive code summaries~\citep{achiam2023gpt, sun2024source}. LLM-powered assistants such as OpenAI's Codex~\citep{chen2021evaluating}, GitHub Copilot~\citep{githubcopilot2021}, and Anthropic's Claude-Code~\citep{anthropic2025claudecode} are increasingly integrated into real-world development workflows to assist engineers in understanding and reviewing code.

However, when the generated summary does not accurately reflect the code's actual implementation, it can cause developers to misunderstand the code, leading to delayed debugging and increased maintenance costs~\citep{8813274, 10.1145/1137983.1138030, IBRAHIM20122293}. Therefore, it is important to accurately and reliably evaluate the factual consistency of LLM-generated code summaries.


The existing reference-based metrics, such as ROUGE~\citep{lin-2004-rouge}, BLEU~\citep{papineni2002bleu}, and METEOR~\citep{banerjee-lavie-2005-meteor}, require human-written reference summaries and measure the lexical overlaps.
However, code summarization is a one-to-many task meaning that semantically accurate summaries may use various wording from the reference~\citep{naik2024crscore, wu2024can}. Consequently, reference-based evaluation methods are limited in detecting factual inaccuracies in such code summaries.

Recent approaches use LLMs as judges, taking the original code and its summary as inputs to assign a consistency score based solely on the LLM's internal knowledge~\citep{zheng2023judging, liu-etal-2023-g, wu2024can}. 
These methods have the advantage of efficiently scoring the code summaries without references or a training process. 
%
%
However, they treat the code summary as a whole and consider factual consistency under a single criterion, producing only a binary or coarse-grained 5-point scale score.
This oversimplified approach has two critical limitations. First, it cannot provide fine-grained assessments of inconsistencies in long summaries. Each sentence may contain different levels or types of factual errors, but identifying these nuances is lost in a single criterion score. Second, they fail to identify which specific sentences are inconsistent or explain the underlying reasons for the inconsistencies, making it difficult to refine and improve the LLM-generated code summary.


Additionally, in real-world code, functions or classes referenced within the input code are often defined externally, and code summaries frequently describe such external elements~\citep{li2024deveval, ding-etal-2024-cocomic, liu2023repobench}. However, existing methods evaluate summaries based solely on the input code without considering this external context, making them unable to accurately assess whether descriptions of externally defined elements are factually consistent. 
For example, as shown in Figure~\ref{figure:intro}, when the summary describes the functionality of `\texttt{key\_prefix}', evaluating its accuracy requires not only the input code but also the external context where it is defined. 


Therefore, we propose a novel evaluation method, \textit{\textbf{ReFEree}}, a reference-free and fine-grained method for evaluating the factual consistency of code summarization in a real-world environment. This method is built upon four representative criteria that should be considered when evaluating real-world code summaries. Based on these criteria, a segment-level evaluation approach is designed to localize factual inconsistencies and provide actionable feedback. Furthermore, a code-related information searching mechanism is employed to accurately evaluate summaries containing external information unavailable in the input code alone.
Our framework combines fine-grained evaluation with explicit code-related evidence, ensuring explainability of how the final consistency score is derived while enabling objective and consistent evaluation by minimizing reliance on internal knowledge.
%

To verify that \textit{ReFEree} is reliable and accurate, we construct an evaluation benchmark that includes human labels of factual consistency for LLM-generated code summaries at both the summary-level and the segment-level. We collect project codes from 125 real-world repositories written in Python and Java.
Using this benchmark, we compare \textit{ReFEree} with 13 existing evaluation methods. The results show that \textit{ReFEree} achieves the highest correlation with human judgment for both languages. Furthermore, various additional experiments demonstrate that our evaluation method can consistently maintain evaluation performance across various environments and serve as a stable and generalizable evaluation.

\begin{figure*}[t]
    \centering
    \includegraphics[width=1\linewidth]{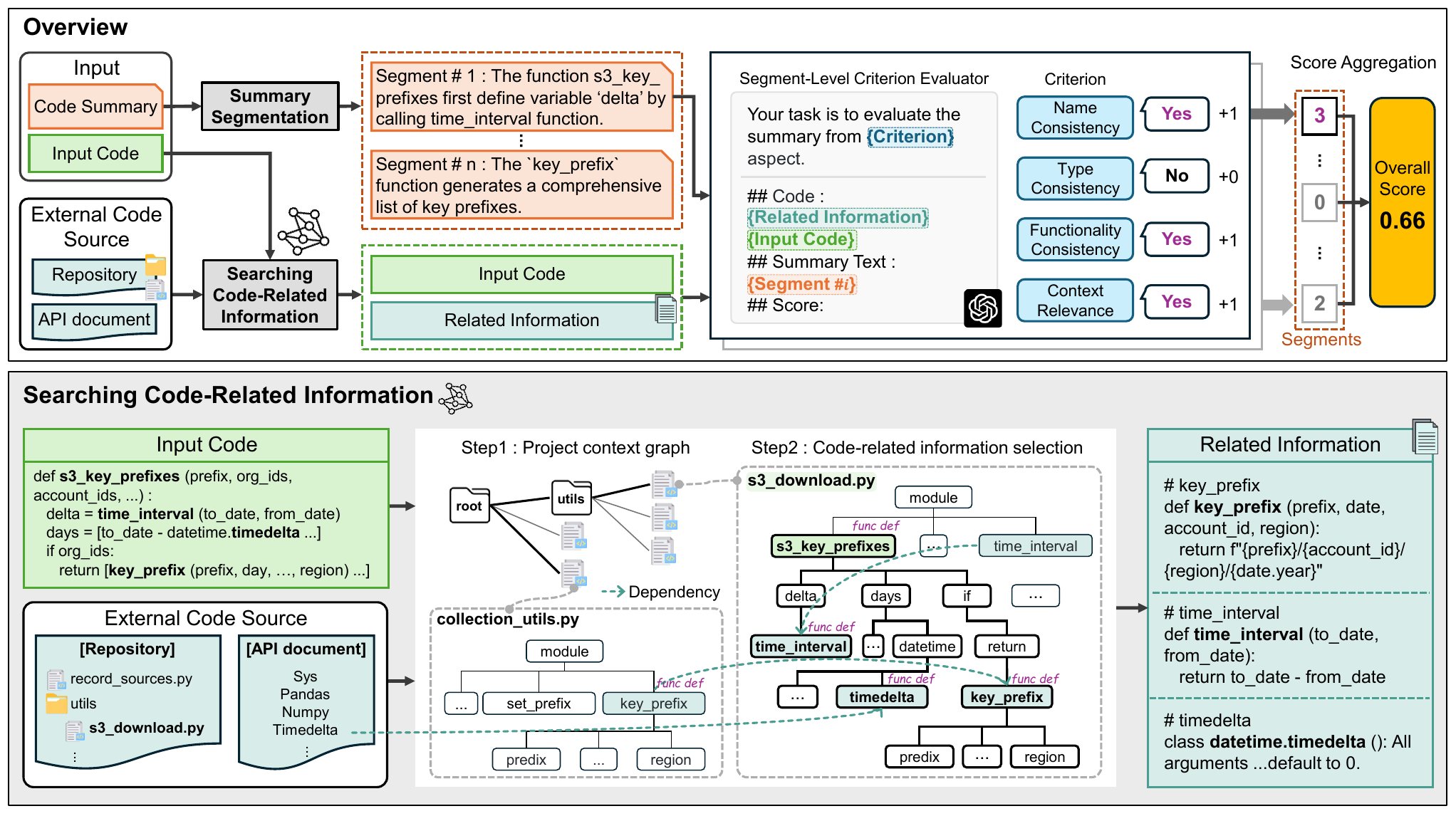}
    \caption{Overview of \textit{ReFEree} framework. Our evaluation process first performs code-related information searching to consider dependency relations in the input code. The summary is then segmented at the sentence level, and an LLM evaluates each segment according to the four factual inconsistency criteria to obtain segment-level scores. These scores are subsequently aggregated to compute the final consistency score.}
    \label{figure:overview}
\end{figure*}


%% file: texs/2_related.tex
\section{Related Works}

\paragraph{Factual Consistency in Code Summarization.}

Factual consistency is essential for evaluating the reliability of LLM-generated code summaries. While there has been considerable research on factual consistency in code generation~\citep{liu2024exploring, tian2025codehalu, zhang2025llm}, those in code summarization remain relatively underexplored.
\citet{kang2024identifying} and \citet{8813274} define inconsistencies in code comments, but only focus on specific aspects such as design constraints and parameter types. \citet{maharaj-etal-2025-etf} propose a cause-oriented taxonomy of error types, but it is not directly applicable to detection. Furthermore, existing studies focus on isolated functions, overlooking inconsistencies involving external dependencies.

\paragraph{Evaluation of Code Summarization.}
Most prior studies have evaluated code summarization using metrics adapted from natural language summarization~\citep{ahmad2020transformer, choi-etal-2021-learning, choi-etal-2023-blocsum}, broadly into two types. First, reference-based methods evaluate textual similarity between human references and generated summaries using n-gram overlap (BLEU~\citep{papineni2002bleu}, ROUGE~\citep{lin-2004-rouge}, METEOR~\citep{banerjee-lavie-2005-meteor}) or embedding similarity (BERTScore~\citep{zhang2019bertscore}, SentenceBERT~\citep{reimers2019sentence}). However, low-quality or outdated references can misrepresent the actual quality of the generated summary.

Second, reference-free methods directly compute similarity between the summary and the code without relying on references. SIDE~\citep{mastropaolo2024evaluating} evaluates the semantic fitness by using contrastive learning to distinguish suitable and unsuitable summaries. However being trained on short summaries, it performs poorly for long, descriptive summaries covering multiple functionalities. 
Inspired by recent NLP advances, several approaches use LLMs as evaluators to simulate human judgment by measuring summary's factuality~\citep{wang-etal-2023-chatgpt, liu-etal-2023-g, min-etal-2023-factscore, song2024finesure}. However, few studies have tailored these methods to the code domain. \citet{wu2024can} propose a judge prompt that assess a code summary as a whole using a single ``factual consistency'' criterion with 5-point scale score, limiting both explainability and granularity. \citet{maharaj-etal-2025-etf} attempt fine-grained evaluation at the entity level, but focus solely on binary inconsistency detection without explaining why an entity is incorrect, and rely on an LLM's internal knowledge without explicitly modeling external dependencies.

%% file: texs/3_method.tex
\section{Methodology}
In this section, we describe our evaluation method. We first explain the process of defining four factual inconsistency criteria (Section~\ref{3.1}). As shown in Figure~\ref{figure:overview}, \textit{ReFEree} proceeds in two stages by first searching input code-related information (Section~\ref{3.2}) and then calculating the overall score via fine-grained evaluation (Section~\ref{3.3}).

\begin{table*}[t]
\centering
\begin{adjustbox}{max width=\textwidth}
\begin{tabular}{cc}
\toprule
\textbf{Criteria (Distribution)} & \textbf{Definition}  \\ \midrule
\begin{minipage}[c]{0.2\linewidth}\centering \textbf{[C1]} Name\\Inconsistency (14\%) \end{minipage} & 
\begin{minipage}[c]{0.95\linewidth}The name of a function, class, or variable in the summary does not match the actual identifier in the input code or mistakenly refers to a different entity with the same name in the project.\end{minipage} \\
\midrule
\begin{minipage}[c]{0.2\linewidth}\centering \textbf{[C2]} Type\\Inconsistency (15\%)\end{minipage} & 
\begin{minipage}[c]{0.95\linewidth}The described function’s return type or variable type in the summary is inconsistent with the actual type in the input code or in directly dependent functions.\end{minipage}  \\
\midrule
\begin{minipage}[c]{0.2\linewidth}\centering \textbf{[C3]} Functionality\\Inconsistency (35\%) \end{minipage} & 
\begin{minipage}[c]{0.95\linewidth}The functionality or purpose described in the summary does not accurately reflect what the input code or its dependent functions actually implement. This often arises when dependency relationships are ignored or misinterpreted, leading to an incorrect description of behavior.\end{minipage} \\
\midrule
\begin{minipage}[c]{0.2\linewidth}\centering \textbf{[C4]} Context\\Irrelevant (33\%) \end{minipage} & 
\begin{minipage}[c]{0.95\linewidth}The summary contains content that is unnecessary or irrelevant to the input code or relevant information, such as descriptions of unrelated entities or overly generalized dependency context that does not contribute to understanding the function.\end{minipage}\\
\bottomrule
\end{tabular}
\end{adjustbox}
\caption{Four factual inconsistency criteria with the proportion of each for evaluating real-world code summarization.}
\label{tab:criteria}
\end{table*}

\subsection{Factual Inconsistency Criteria} \label{3.1}
To systematically evaluate the factual consistency of real-world code summaries, well-defined and generalizable criteria are required. Since LLM-generated summaries exhibit distinct error patterns, often stemming from hallucination or misinterpretation of code semantics, we empirically analyze the patterns of factual inconsistencies specific to LLM-generated summaries and identify four representative patterns to serve as evaluation criteria.
We randomly sample 100 real-world Python and Java functions containing entities with dependency chains from popular repositories and generate descriptive summaries using three LLMs, Qwen-Coder, CodeLlama, and GPT-4 to ensure that the identified patterns are generalizable and not artifacts of a specific model's behavior.
Three human annotators then review all 300 summaries and manually annotate any factual inconsistencies, with their comments (e.g., type errors, functionality inaccuracies).

Our analysis reveals four main patterns of factual inconsistencies in LLM-generated code summaries. The most frequent is functionality inconsistency (35\%), where explanations diverge from actual implementations due to misinterpreted dependency relations. The second most common is irrelevant context (33\%), where the generated summary introduces functionality, purpose, or usage context that cannot be verified from the code itself and thus appears to be hallucinated. Unlike functionality inconsistency, which reflects a misreading of existing code behavior, irrelevant context represents a more deceptive form of factual error: ungrounded claims are presented as facts, directly distorting the developer's perception of what the code actually does.

The remaining patterns are name (14\%) and type inconsistency (15\%), involving incorrect descriptions of code element names or return types. The remaining 3\% consist of minor issues, such as content duplication, that do not critically hinder code understanding.

Based on this analysis, we define four evaluation criteria specifically tailored for assessing the factual consistency of LLM-generated real-world code summaries. Table~\ref{tab:criteria} presents these criteria along with their definitions. More detailed explanations and examples can be found in Appendix~\ref{A.1}.

\subsection{Searching Code-Related Information} \label{3.2}
Real-world code summaries generated by LLMs often describe external elements such as function calls, class hierarchies, and API usage. Therefore, we adopt a two-step approach based on static program analysis approach to provide objective evidence for evaluating such external descriptions. Further details are provided in Appendix~\ref{A.2}.

\paragraph{Step 1: Project context graph construction.}
We first construct a project context graph. Given a project repository that contains the input code targeted for evaluation, we traverse the abstract syntax tree (AST)~\citep{8812062} of each file in the repository to collect the code entities. Based on the parsed information, we construct a heterogeneous directed acyclic graph, where all types of code entities within the project are represented as nodes, and their relationships are linked as directed edges. 
Each node stores metadata such as the entity’s AST type and code context, and each edge represents dependency relationships between two nodes, such as assignment, class inheritance, or function calls.

\paragraph{Step 2: Code-related information selection.}
Including all project context can produce noise that is not essential for understanding the core functionality~\citep{lomshakov-etal-2024-proconsul}. To mitigate the interference of trivial or unnecessary relations in summary evaluation, we adopt \textit{crucial entity selection} and a \textit{1-hop dependency searching} strategy to selectively search only the dependency information that is essential and directly relevant to code summary evaluation.

We begin by performing a depth-first search (DFS) on the project context graph, starting from each entity node in the input code to traverse all related nodes connected via dependency edges. Among all related nodes, we first selectively search entities corresponding to `functions', `classes', and `variables' that play a central role in real-world code understanding, as empirically determined through experiments at previous research~\citep{cheng-etal-2024-dataflow, luo-etal-2024-repoagent}.
Second, instead of exploring all dependency relations in the input code, we stop further searches once the dependency context of each core entity has been searched. This approach is based on prior research showing that multi-hop searches in hierarchical dependencies can increase noise for summary understanding as the number of hops increases~\citep{zhang2025llm}. If an entity has external dependencies, we retrieve information from predefined API documentation (Ablation analysis comparing different n-hop search configurations is provided in Appendix~\ref{5.6}).




\subsection{Calculating Factual Consistency Score} \label{3.3}

For fine-grained evaluation, we first apply the NLTK sentence tokenizer~\footnote{\url{https://www.nltk.org/api/nltk.tokenize.html}} to split a code summary ($\mathcal{D}$) into sentence-level segments, resulting in a set of segments $\mathcal{D}=\{S_1, S_2, ... , S_i, ...,S_n\}$.

We then evaluate each segment against the four criteria defined in Section~\ref{3.1}. For evaluation, we construct a criterion evaluation prompt as briefly illustrated in Figure~\ref{figure:overview} (complete prompt in Appendix~\ref{A.3}), and use an LLM to determine whether each segment contains factual inconsistencies with respect to each criterion, considering both the input code and related information obtained in Section~\ref{3.2}. 
The evaluator $f(S,C)$ outputs 0 if any factual inconsistency is detected in segment $S$ with respect to criterion $C$, and 1 otherwise.

The total segment-level score is calculated by summing the outputs across all four criteria. Then, the segment-level scores are aggregated to calculate the final overall score. Given a summary segmented as $\mathcal{D}$ and a set of criteria $Criteria=\{C_1, C_2, C_3, C_4\}$, the overall factual consistency score is computed as follows:

{\small
\[
SCORE=\frac{1}{|\mathcal{D}|\times|Criteria|} \sum_{S \in \mathcal{D}} \sum_{C\in Criteria} f(S, C)
\]}
\noindent The score ranges from 0 to 1, with higher values indicating greater factual consistency. This decompose-and-aggregate approach enables fine-grained assessment while ensuring interpretability of the overall scoring process.


%% file: texs/4_experiment_setups.tex
\section{Experiment Setups}

\subsection{Evaluation Benchmark} \label{4.1}
To verify the effectiveness of our proposed evaluation method, it is essential to demonstrate that its evaluation results using our method are highly correlated with human judgment. However, no existing benchmark provides human-annotated factual consistency labels for LLM-generated code summaries. Moreover, naturally generated LLM summaries lack sufficient hallucinated examples to rigorously test inconsistency detection capabilities. Therefore, we construct a reliable set of human labels for factual consistency at both the summary and segment levels. In this section, we briefly describe the labeling process in three steps; more detailed procedures, further analysis, statistics, and examples are provided in Appendix~\ref{B.1}. 


\paragraph{Step 1: Source code collection.}
We build our benchmark by extracting project codes from real-world open-source repositories: 1,825 Python functions from DevEval~\citep{li2024deveval} (115 repositories) and 230 Java functions from ClassEval~\citep{du2023classeval} (10 repositories). Approximately 86\% of the functions are non-standalone, containing context-aware dependencies. 
Both languages are prevalent in open-source environments and feature diverse dependency relationships, ensuring a wide range of code complexity. 

\paragraph{Step 2: Summary generation.}
We use ChatGPT to generate long and descriptive code summaries, intentionally including at least one factually inconsistent sentence in each summary. 
We design a structured instruction prompt consisting of three components: an instruction description, a set of inconsistency patterns with demonstrations~\footnote{These patterns are derived from empirically observed error patterns identified in real-world summary generation scenarios, as analyzed in Section~\ref{3.1}. Therefore, our generated summaries are designed to reflect the real-world distribution of mistakes that naturally occur.}, and the target code along with related information, intentionally guiding the LLM to generate summaries that include inconsistency contents arising when such related information is not properly considered.



\begin{table*}[t]
\begin{adjustbox}{max width=\textwidth}
\begin{tabular}{lcccccccc}
\toprule
\multicolumn{1}{c|}{}                         & \multicolumn{4}{c|}{\textbf{Python}}                                                                                 & \multicolumn{4}{c}{\textbf{Java}}                                                               \\ \cline{2-9} 
\multicolumn{1}{c|}{\multirow{-2}{*}{\textbf{Methods}}} & \textbf{$r_p$}        & \textbf{$r_s$}       & \multicolumn{1}{c|}{\textbf{$\tau$}}        & \multicolumn{1}{c|}{\textbf{Average}}        & $r_p$        & $r_s$       & \multicolumn{1}{c|}{$\tau$}         & \textbf{Average}        \\ \midrule
\multicolumn{9}{c}{\cellcolor[HTML]{EDEDED}\textit{reference-based methods}}                                                                                                                                                                         \\ \midrule
\multicolumn{1}{l|}{ROUGE-1~\citep{lin-2004-rouge}}                  & 0.046          & 0.044          & \multicolumn{1}{c|}{0.034}          & \multicolumn{1}{c|}{0.041}          & 0.201          & 0.186          & \multicolumn{1}{c|}{0.140}          & 0.176          \\
\multicolumn{1}{l|}{ROUGE-2~\citep{lin-2004-rouge}}                  & 0.030          & 0.031          & \multicolumn{1}{c|}{0.024}          & \multicolumn{1}{c|}{0.028}          & 0.194          & 0.176          & \multicolumn{1}{c|}{0.137}          & 0.169          \\
\multicolumn{1}{l|}{ROUGE-L~\citep{lin-2004-rouge}}                  & 0.045          & 0.037          & \multicolumn{1}{c|}{0.029}          & \multicolumn{1}{c|}{0.037}          & 0.202          & 0.178          & \multicolumn{1}{c|}{0.137}          & 0.172          \\
\multicolumn{1}{l|}{BLEU~\citep{papineni2002bleu}}                     & 0.016          & 0.018          & \multicolumn{1}{c|}{0.015}          & \multicolumn{1}{c|}{0.017}          & 0.081          & 0.065          & \multicolumn{1}{c|}{0.058}          & 0.068          \\
\multicolumn{1}{l|}{METEOR~\citep{banerjee-lavie-2005-meteor}}                   & 0.033          & 0.024          & \multicolumn{1}{c|}{0.019}          & \multicolumn{1}{c|}{0.025}          & 0.135          & 0.139          & \multicolumn{1}{c|}{0.105}          & 0.126          \\
\multicolumn{1}{l|}{BERTScore~\citep{zhang2019bertscore}}                & 0.015          & 0.000          & \multicolumn{1}{c|}{0.000}          & \multicolumn{1}{c|}{0.005}          & 0.189          & 0.149          & \multicolumn{1}{c|}{0.113}          & 0.150          \\
\multicolumn{1}{l|}{SBCS~\citep{reimers2019sentence}}                     & 0.068          & 0.036          & \multicolumn{1}{c|}{0.028}          & \multicolumn{1}{c|}{0.044}          & 0.363          & 0.227          & \multicolumn{1}{c|}{0.178}          & 0.256          \\
\multicolumn{1}{l|}{SBED~\citep{reimers2019sentence}}                     & -0.058         & -0.036         & \multicolumn{1}{c|}{-0.028}         & \multicolumn{1}{c|}{-0.041}         & -0.348         & -0.227         & \multicolumn{1}{c|}{-0.178}         & -0.251         \\ \hline
\multicolumn{9}{c}{\cellcolor[HTML]{EDEDED}\textit{reference-free methods}}                                                                                                                                                                          \\ \midrule
\multicolumn{1}{l|}{SIDE~\citep{mastropaolo2024evaluating}}                     & -0.064         & -0.056         & \multicolumn{1}{c|}{-0.043}         & \multicolumn{1}{c|}{-0.054}         & 0.032          & 0.024          & \multicolumn{1}{c|}{0.019}          & 0.025          \\
\multicolumn{1}{l|}{LLM-judge~\citep{zheng2023judging}}                & 0.419          & 0.410          & \multicolumn{1}{c|}{0.360}          & \multicolumn{1}{c|}{0.396}          & 0.385          & 0.363          & \multicolumn{1}{c|}{0.318}          & 0.355          \\
\multicolumn{1}{l|}{G-Eval~\citep{liu-etal-2023-g}}                   & 0.427          & 0.413          & \multicolumn{1}{c|}{0.360}          & \multicolumn{1}{c|}{0.400}          & 0.457          & 0.407          & \multicolumn{1}{c|}{0.355}          & 0.406          \\
\multicolumn{1}{l|}{Factscore~\citep{min-etal-2023-factscore}}                & 0.410          & 0.426          & \multicolumn{1}{c|}{0.338}          & \multicolumn{1}{c|}{0.391}          & 0.391          & 0.365          & \multicolumn{1}{c|}{0.310}          & 0.355          \\
\multicolumn{1}{l|}{CODERPE~\citep{wu2024can}}                  & 0.418          & 0.405          & \multicolumn{1}{c|}{0.353}          & \multicolumn{1}{c|}{0.392}          & 0.457          & 0.398          & \multicolumn{1}{c|}{0.347}          & 0.401          \\ \midrule
\multicolumn{1}{l|}{\textit{ReFEree} (w/o info)}  & 0.432 & 0.432 & \multicolumn{1}{c|}{0.349} & \multicolumn{1}{c|}{0.404} & 0.469 & 0.458 & \multicolumn{1}{c|}{0.387} & 0.438 \\
\multicolumn{1}{l|}{\textbf{\textit{ReFEree} (w/ info)}}  & \textbf{0.497} & \textbf{0.489} & \multicolumn{1}{c|}{\textbf{0.390}} & \multicolumn{1}{c|}{\textbf{0.459}} & \textbf{0.515} & \textbf{0.502} & \multicolumn{1}{c|}{\textbf{0.423}} & \textbf{0.480} \\ \bottomrule
\end{tabular}
\end{adjustbox}
\caption{We compare our method (\textit{ReFEree}) with 8 reference-based and 5 reference-free methods using Pearson ($r_p$), Spearman ($r_s$), and Kendall’s Tau ($\tau$) correlation coefficients at the summary level. Our method achieves \textit{p}-values all below 0.005, indicating statistical significance. The best result is shown in \textbf{bold}.}
\label{tab:main1}
\end{table*}

\paragraph{Step 3: Factual consistency labeling.}
We then annotate factual consistency ground truth labels at both summary and segment levels. At the summary level, each summary is labeled on a 5-point scale (1: highly inconsistent to 5: highly consistent). At the segment level, each sentence is labeled as \texttt{CORRESPOND} (1) or \texttt{NOT CORRESPOND} (0) to each of the four criteria (C1–C4) defined in Section~\ref{3.1}.

To obtain reliable and accurate labels, we adopt a Human–AI collaborative labeling approach, which has proven effective in producing high-quality annotations~\citep{li-etal-2023-halueval, zhang2023peanut, kim2024meganno+}. First, three LLMs with different roles determine the candidate label through majority voting. 
If all three LLMs may assign different scores during summary-level labeling (since there are five possible labels), the score closest to the mean is selected as the candidate label.
Then, three human annotators subsequently post-edit the candidate label. The final label is determined according to the following three rules: 
(1) If at least two annotators agree with the candidate label, the label is retained.
(2) If at least two annotators revise to the same label, the label is edited accordingly.
(3) For summary-level labeling: If all annotators assign different labels, the final label is determined through further discussion among the three annotators.

We compute Krippendorff’s alpha reliability~\citep{krippendorff2018content} to assess annotator agreement. We achieve an agreement of 0.74 at the summary level and an average of 0.84 at the segment level, both indicating a high level of agreement. All three human annotators hold at least a Master's degree in Computer Science and have an average of over four years of experience in Python and Java programming, ensuring the expertise and credibility of the annotations. 

\subsection{Baselines} \label{4.2}
We select 13 commonly used evaluation methods, 8 reference-based and 5 reference-free methods, from code summarization tasks to compare with our proposed method. Detailed explanations are provided in Appendix~\ref{B.2}.

\subsection{Implementation Details} \label{4.3}
Our method supports various LLMs, including both closed and open-source models, as segment-level criterion evaluators. In our main experiments, we use OpenAI’s GPT-4.1-mini to ensure a fair comparison with existing baselines under the same setting. All evaluations are conducted with consistent hyperparameters: temperature = 0.1, top-p = 0.9, top-k = 50, and max new tokens = 4. All experiments are conducted three times with different seeds, and the average results are reported. Evaluating the code summary using \textit{ReFEree} incurs a cost of only \$0.004 per sample. A detailed comparison of time and cost with existing methods is provided in Appendix~\ref{C.4}.

%% file: texs/5_result_analysis_v2.tex
\section{Results and Analyses}

\begin{figure}[t]
    \centering
    \includegraphics[width=1\linewidth]{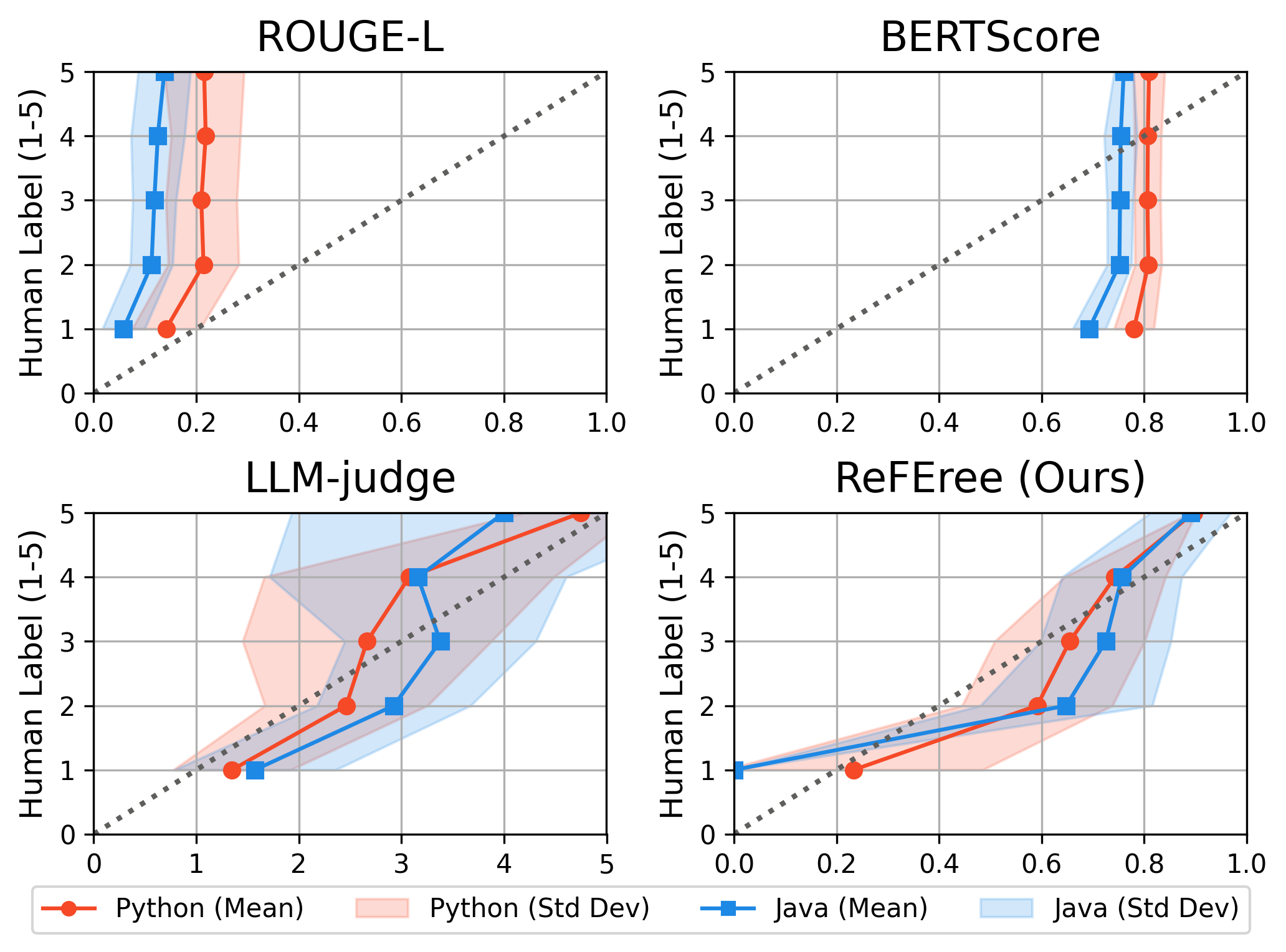}
    \vspace{-0.6cm}
    \caption{Correlation between human judgment and the scores from each method. Each plot shows the relationship between scores obtained from each method \textbf{(x-axis) }and human labels \textbf{(y-axis)}, with \textcolor{red}{red} for Python and \textcolor{blue}{blue} for Java. The \textcolor{gray}{gray} dashed line represents perfect correlation. Since human labels range from 1 to 5, there are no values at y = 0.}
    \label{figure:compare_graph}
\end{figure}

\subsection{Comparisons with Baselines} \label{5.1}

To verify whether our method exhibits a high correlation with human judgment, we compute the Pearson ($r_p$), Spearman ($r_s$), and Kendall's tau ($\tau$) correlation coefficient between summary-level factual consistency scores by humans and those produced by each method. Table~\ref{tab:main1} presents the main results of this comparison.

\textit{ReFEree} outperforms all baselines. \textit{ReFEree} achieves the highest correlation with human judgment in both Python and Java environments. Compared to G-Eval, the strongest baseline, \textit{ReFEree} achieves approximately 15\% (Python) and 18\% (Java) higher correlation.
FactScore decomposes summaries into atomic facts but evaluates each with only a single consistency criterion. \textit{ReFEree}, by contrast, outperforms FactScore by 17.4\% (Python) and 35.2\% (Java) through multi-criteria segment-level evaluation specifically designed for real-world code summaries.

We further conduct an ablation study to assess the impact of code-related information. When related information is excluded (\textit{ReFEree} (w/o info)), correlation with human judgment decreases by approximately 0.05 in both Python and Java. This demonstrates that our information searching module plays a vital role in achieving accurate factual consistency evaluation.



\begin{table}[t]
\centering
\begin{adjustbox}{max width=\linewidth}
\begin{tabular}{l|cccc|c}
\toprule
\textbf{Criteria} & \textbf{C1} & \textbf{C2} & \textbf{C3} & \textbf{C4} & \textbf{Average} \\ \midrule
Python & 0.922 & 0.961 & 0.931 & 0.924 & 0.934 \\
Java & 0.913 &	0.991 &	0.940 &	0.876 &	0.930 \\
\bottomrule
\end{tabular}
\end{adjustbox}
\caption{Accuracy of segment-level factual inconsistency evaluation. We report the accuracy of \textit{ReFEree} in identifying four criteria (C1–C4) defined in Section~\ref{3.1}.}
\label{tab:main2}
\end{table}

To further analyze how each method evaluates factual consistency in comparison to human judgment, we compare the score distributions of ROUGE-L, BERTScore, LLM-judge, and \textit{ReFEree}. Figure~\ref{figure:compare_graph} visualizes the relationship between evaluation scores and human labels by plotting the mean and standard deviation of scores corresponding to each human label.
The observed patterns reflect inherent characteristics of each method. ROUGE-L consistently assigns low scores even to factually consistent summaries due to its reliance on lexical overlap, while BERTScore tends to assign high scores even to summaries with inaccurate descriptions. LLM-judge show moderate correlation with human judgment but exhibit high variance, indicating lower consistency compared to our method. In contrast, \textit{ReFEree} demonstrates strong and stable alignment with human judgment, with most data points closely following the ideal correlation line.

%
%
%


\subsection{Performance of Segment Level Evaluator}  \label{5.2}
To verify that our segment-level evaluator can accurately identify factual inconsistencies and their types, we analyze prediction accuracy in Table~\ref{tab:main2}. The results show that \textit{ReFEree} achieves an average accuracy of 0.934 (Python) and 0.930 (Java), demonstrating reliable detection of segment-level factual inconsistencies. Furthermore, Table~\ref{tab:example_1} provides qualitative comparisons with other methods, illustrating how \textit{ReFEree} offers explainability by pinpointing where inconsistencies occur and why they are classified as such.
Also in Appendix~\ref{C.3}, we briefly demonstrate that \textit{ReFEree} can also be applied to a real-world developer-written summary.

\begin{table}[t]
\centering
\begin{adjustbox}{max width=\linewidth}
\begin{tabular}{l}
\toprule
\begin{minipage}[t]{1.3\linewidth}
\textbf{\#\# Code \#\#} \\
def convert\_to\_idn(url):\\
\hspace*{1em}parts = list(urllib.parse.urlsplit(url))\\
\hspace*{1em}try:\\
\hspace*{2em}parts{[}1{]}.encode(`ascii')\\
\hspace*{1em}except UnicodeEncodeError:\\
\hspace*{2em}\textcolor{teal}{// Some code $\dotsc$} \\
\hspace*{2em}if port:\\
\hspace*{3em}parts{[}1{]} += `:' + port\\
\hspace*{3em}return urllib.parse.urlunsplit(parts)\\
\hspace*{2em}else:\\
\hspace*{3em}return url \end{minipage} \\ \midrule
\textbf{Human Label : 0.60} (Normalized) \\ \midrule
ROUGE-L : 0.18 \hspace*{1em} BERTScore : 0.81 \\
SIDE : 0.85 \hspace*{3.2em} G-Eval : 0.40 (Normalized) \\ \midrule
\textbf{\textit{ReFEree} score} : \textbf{0.55} \\
\begin{minipage}[t]{1.3\linewidth}
\textbf{Summary} : The `convert\_to\_idn` function is designed to convert the hostname portion of a URL into its Internationalized Domain Name (IDN) ASCII-compatible encoding using the Punycode standard. \textcolor{violet}{It first splits the URL into its components using the `urlparse` function , then attempts to encode the hostname in ASCII \textbf{(C1, C3)}.} [Some explanations $\dotsc$] \textcolor{violet}{The function returns a list of URL components with the converted hostname \textbf{(C2, C3)}}. \textcolor{violet}{This utility is particularly useful in web crawlers and multilingual domain management systems where URLs need to be normalized for consistent processing and storage \textbf{(C4)}.}\end{minipage} \\ \bottomrule
\end{tabular}
\end{adjustbox}
\caption{Qualitative comparison of \textit{ReFEree} and existing methods. Factually inconsistent segments are highlighted in \textcolor{violet}{purple}, with C1–C4 indicating the violated criteria (defined in Table~\ref{tab:criteria}). Human label and G-Eval scores are normalized to 0\textasciitilde1 for comparison. More comparisons are described in Appendix~\ref{C.5}.}
\label{tab:example_1}
\end{table}

\begin{table}[t]
\centering
\begin{adjustbox}{max width=\linewidth}
\begin{tabular}{lcc}
\hline
\multicolumn{1}{c}{Methods}      & \textbf{Python}         & \textbf{Java}           \\ \hline
G-Eval                    & 0.400    & 0.406   \\ \hline
ReFEree (w/ C1)           & 0.394 (\textcolor{red}{-1.6\%})  & 0.318 (\textcolor{red}{-5.2\%})  \\
ReFEree (w/ C2)           & 0.417 (\textcolor{blue}{+4.3\%}) & 0.379 (\textcolor{red}{-6.8\%})   \\
ReFEree (w/ C3)           & 0.419 (\textcolor{blue}{+4.8\%})   & 0.391 (\textcolor{blue}{+12.7\%})   \\
ReFEree (w/ C4)           & 0.405 (\textcolor{blue}{+1.2\%})   & 0.352 (\textcolor{blue}{+3.1\%})  \\ \hline
\textbf{ReFEree (w/ all)} & \textbf{0.459} (\textcolor{blue}{+14.6\%}) & \textbf{0.480} (\textcolor{blue}{+18.1\%}) \\ \hline
\end{tabular}
\end{adjustbox}
\caption{Average correlation coefficients (Pearson, Spearman, Kendall's Tau) for \textit{ReFEree} (w/ all), its single-criterion ablations (w/C1-C4), and G-Eval. Percentages indicate relative performance change compared to G-Eval.}
\label{tab:experient_criteria}
\end{table}

\subsection{Effectiveness of Considering Four Criteria}  \label{5.3}
To demonstrate that considering all four criteria when evaluating a segment-level code summary is effective, we compare \textit{ReFEree} using all four criteria (w/ all) against single-criterion ablations (w/ C1–C4) in Table~\ref{tab:experient_criteria}. We also report performance changes relative to the previous SOTA, G-Eval.

Experimental results show that evaluating all four inconsistency criteria in real-world code summaries produces the most reliable assessment. Among the individual criteria, considering only functionality inconsistency (w/ C3) results in the smallest performance drop, while considering only name inconsistency (w/ C1) causes the largest drop, even falling below G-Eval. These findings suggest that name inconsistencies (C1) are less influential when humans judge the factual consistency of code summaries, whereas functionality (C3) and type inconsistencies (C2) play a more critical role, highlighting their importance in human evaluations. We also analyze how assigning different weights to each criterion during score aggregation affects evaluation performance in Appendix~\ref{C.2}.

\begin{table}[t]
\centering
\begin{adjustbox}{max width=\linewidth}
\begin{tabular}{clll}
\hline
\textbf{Models}                     & \multicolumn{1}{c}{\textbf{Methods}} & \multicolumn{1}{c}{\textbf{Python}} & \multicolumn{1}{c}{\textbf{Java}} \\ \hline
\multirow{3}{*}{Llama3 (8B)}        & G-Eval                     & 0.145                      & 0.217                    \\
                                    & G-Eval (w/ info)            & 0.171                      & 0.251                    \\
                                    & \textbf{ReFEree}           & \textbf{0.187}             & \textbf{0.367}           \\ \hline
\multirow{3}{*}{Mistral (7B)}       & G-Eval                     & 0.191                      & 0.196                    \\
                                    & G-Eval (w/ info)            & 0.191                      & 0.242                    \\
                                    & \textbf{ReFEree}           & \textbf{0.233}             & \textbf{0.250}           \\ \hline
\multirow{3}{*}{Qwen2 (7B)}         & G-Eval                     & 0.221                      & 0.185                    \\
                                    & G-Eval (w/ info)            & 0.243                      & 0.238                    \\
                                    & \textbf{ReFEree}           & \textbf{0.284}             & \textbf{0.304}           \\ \hline
\multirow{3}{*}{Qwen2.5-Coder (7B)} & G-Eval                     & 0.247                      & 0.243                    \\
                                    & G-Eval (w/ info)            & 0.304                      & 0.348                    \\
                                    & \textbf{ReFEree}           & \textbf{0.351}             & \textbf{0.394}           \\ \hline
\multirow{3}{*}{GPT4o-mini}         & G-Eval                     & 0.242                      & 0.390                    \\
                                    & G-Eval (w/ info)            & 0.259                      & 0.428                    \\
                                    & \textbf{ReFEree}           & \textbf{0.354}             & \textbf{0.429}           \\ \hline
\multirow{3}{*}{GPT4.1-mini}        & G-Eval                     & 0.400                      & 0.406                    \\
                                    & G-Eval (w/ info)            & 0.418                      & 0.463                    \\
                                    & \textbf{ReFEree}           & \textbf{0.459}             & \textbf{0.480}           \\ \hline
\end{tabular}
\end{adjustbox}
\caption{Generalizability of \textit{ReFEree} Across LLMs. We compare the performance improvements of \textit{ReFEree} across six different LLMs, relative to G-Eval and G-Eval (w/ info) for the average of Pearson, Spearman, and Kendall’s Tau correlation coefficients ($p$ $< 0.005$).
}
\label{tab:main5}
\end{table}

\subsection{Generalizability Across Different LLMs}  \label{5.4}

To demonstrate that \textit{ReFEree} consistently outperforms existing methods across different LLMs, we conduct a comparative analysis against the G-Eval (previous SOTA). As shown in Table~\ref{tab:main5}, we compare the human correlations when using six various LLMs, respectively, as our segment-level criterion evaluator, four open-LLMs (Llama3, Mistral, Qwen2, and Qwen2.5 coder), and two closed-LLMs (GPT4o-mini and GPT4.1-mini).

As discussed in Section~\ref{5.1}, \textit{ReFEree} using GPT4.1-mini as the evaluator achieves about a 15\% improvement over G-Eval, and similar improvements are observed when using other LLMs. 
While open small models show lower correlations than closed LLMs, our method still outperforms G-Eval, also when G-Eval incorporates both input code and related information (G-Eval w/ info), demonstrating that our fine-grained, systematic evaluation process achieves consistent performance improvements across different LLM evaluator settings.

\begin{table}[t]
\centering
\begin{adjustbox}{max width=\linewidth}
\begin{tabular}{c|cc|cc}
\hline
\textbf{Settings} & \multicolumn{2}{c|}{\textbf{Different Prompts}} & \multicolumn{2}{c}{\textbf{Different Seeds}} \\ \hline
Methods           & G-Eval             & ReFEree                    & G-Eval            & ReFEree                  \\ \hline
Python            & 0.683              & \textbf{0.733}             & 0.805             & \textbf{0.893}           \\
Java              & 0.400              & \textbf{0.730}             & 0.712             & \textbf{0.901}           \\ \hline
\end{tabular}
\end{adjustbox}
\caption{Stability of \textit{ReFEree}.
We compute inter-annotator agreement (IAA) scores across three independent evaluation results for each method (G-Eval and ReFEree) when evaluated under different prompts or different seed settings, for both G-Eval and ReFEree.}
\label{tab:stability}
\end{table}


\subsection{Stability in Evaluation Results}  \label{5.5}
Stability is a critical aspect of any evaluation method, as consistent assessment is essential. From this perspective, Table~\ref{tab:stability} evaluates the stability of \textit{ReFEree} compared to G-Eval by measuring inter-annotator agreement (IAA) across three repeated evaluations under different seeds and prompts (prompt variations detailed in Appendix~\ref{C.1}).
Under seed variation, \textit{ReFEree} achieves IAA of 0.893 (Python) and 0.901 (Java), demonstrating highly consistent assessments across runs. Under prompt variation, G-Eval shows significant instability with IAA of 0.683 (Python) and 0.400 (Java), whereas \textit{ReFEree} maintains higher IAA of 0.733 (Python) and 0.730 (Java). These results indicate that \textit{ReFEree}'s fine-grained evaluation approach provides more stable and reliable assessments than conventional LLM-based methods.

%% file: texs/6_conclusion.tex
\section{Conclusions}
We introduced a novel reference-free and fine-grained method for assessing the factual consistency of real-world code summarization.
We also constructed a high-quality human-labeled benchmark to validate the reliability of the proposed method. 
The comprehensive evaluation results demonstrate that ReFEree achieves high accuracy, reliability, generalizability, and stability, making it effectively applicable to real-world code summarization tasks.

%

\section*{Limitations}

\textit{ReFEree} focuses specifically on evaluating whether LLM-generated summaries contain factually inconsistent information with the code, rather than assessing completeness or overall quality. This design choice is intentional. As highlighted in prior work~\citep{wang2020asking, min-etal-2023-factscore}, factual consistency is a key factor in ensuring basic reliability and has been repeatedly emphasized as a critical problem that warrants dedicated attention. A factually inconsistent summary can mislead developers regardless of how comprehensive it is. Extending to evaluate completeness or overall summary quality is a promising direction for future work. Also, our approach relies on AST-based static analysis and does not handle dynamic language features (e.g., dynamic dispatch in Python). However, such features accounted for only 0.07\% of entities in our benchmark, with no observable performance degradation, suggesting that static syntactic dependencies suffice for reliable factual verification in practice.



%% file: texs/7_appendix_v2.tex
\clearpage
\appendix

\section{Additional Details About ReFEree} \label{A}


\subsection{More details of Factual Inconsistency Criteria Analysis}  \label{A.1}
In Section~\ref{3.1}, we conduct an analysis to accurately identify factual inconsistency patterns that occur in real-world code summaries. To this end, we generate a total of 300 code summaries using three different models: Qwen2.5-Coder (7B), CodeLlama (7B), and GPT-4 (gpt-4o-mini). For summary generation, we use identical hyperparameters across all models: temperature = 0.6, top-p = 0.9, and max new tokens = 256. The prompt template used for generation is shown below, where \textbf{\texttt{\{lang\}}} corresponds to either Python or Java.

\begin{tcolorbox}[
  title=Code summarization prompt,
  colback=gray!15,
  colframe=gray!60,
  coltitle=black,
  fonttitle=\bfseries,
  boxrule=0.1pt,
  boxsep= 2pt,
  left=3pt,
  right=3pt,
  top=2pt,
  bottom=2pt
]
You are an expert in understanding \textbf{\texttt{\{lang\}}} code. As a \textbf{\texttt{\{lang\}}} expert, please generate a detailed and informative summary of the following python function. The generated summary should describe the implementation details of any methods that the function use it. \\
Provide ONLY the summary texts. Do not include any other codes, or notes.
\\
\#\# CODE: \textcolor{blue}{\{input\_code\}} \\
\#\# SUMMARY:
\end{tcolorbox}

Three human annotators independently review each generated summary and manually annotate any parts of factual inconsistencies along with their comments. To ensure annotation reliability, we only count an inconsistency when at least two out of three annotators identify the same comment in the same segment.
Table~\ref{tab:empirical_analysis_v2} presents the frequency of each factual inconsistency type across different models. Our analysis reveals that functionality inconsistency (C3) and context irrelevant (C4) are the most frequently occurring error types across all models, followed by name inconsistency (C1) and type inconsistency (C2). The remaining 3\% of cases, categorized as Others, mostly consist of repeated or redundant context within the summary.
Additionally, Table~\ref{tab:cases_example} provides representative examples of each factual inconsistency criterion identified through our human analysis.

\begin{table}[t]
\centering
\begin{adjustbox}{max width=\linewidth}
\begin{tabular}{c|ccccc|c}
\toprule
\textbf{Pattern} & \textbf{C1} & \textbf{C2} & \textbf{C3} & \textbf{C4} & \textbf{Others} & \textbf{Sum} \\ \midrule
Qwen2.5-Coder & 14 & 24 & 51 & 39 & 4 & 132 \\ 
CodeLlama & 23 & 24 & 55 & 52 & 5 & 159 \\ 
GPT4 & 15 & 10 & 28 & 36 & 4 & 93 \\ \midrule
Sum & 52 & 58 & 134 & 127 & 13 & 384 \\ \midrule
Distribution & 14\% & 15\% & 35\% & 33\% & 3\% & 100\% \\ \bottomrule
\end{tabular}
\end{adjustbox}
\caption{The frequency of factual inconsistency patterns for each model, and the overall distribution of these patterns. The four patterns (C1-C4) are followed in Section~\ref{3.1}.}
\label{tab:empirical_analysis_v2}
\end{table}

\begin{table*}[t]
\centering
\begin{adjustbox}{max width=\textwidth}
\begin{tabular}{cccc}
\toprule
\textbf{Criteria} & \textbf{Input Code} & \textbf{Generated Summary} & \textbf{Inconsistency Reason} \\ \midrule
\begin{minipage}[t]{0.17\linewidth}\centering \textbf{C1.} Name\\Inconsistency\end{minipage} & 
\begin{minipage}[t]{0.5\linewidth}
def where\_filters(request, database, datasette): \\
\hspace*{1em}\textcolor{teal}{// Some code $\dotsc$} \\
\hspace*{1em}extra\_wheres\_for\_ui = [ \\
\hspace*{2em}\{ \\
\hspace*{3em}"text": text, \\
\hspace*{3em}"remove\_url": \textcolor{red}{path\_with\_removed\_args \\
\hspace*{9em}(request, {"\_where": text})},\\
\hspace*{2em}\} for text in request.getlist("\_where") ]\\
\hspace*{1em}\textcolor{teal}{// Some code $\dotsc$} \\
\hspace*{1em}return inner\\
\end{minipage} &
\begin{minipage}[t]{0.25\linewidth}
[...] This removal URL is constructed using the helper function `path\_with\_removed\_args', which builds a new URL path by excluding query parameters identified by the \textcolor{red}{`should\_exclude'} function. [...]
\end{minipage} & \begin{minipage}[t]{0.4\linewidth}
Generated summary describes that the sentence explaining the functionality states that the `should\_exclude' function inside `path\_with\_removed\_args' identifies the query parameters to be excluded. \textbf{However, in the actual code, the function is defined as `should\_remove', not `should\_exclude'}. Considering the related dependencies information within the input code, this discrepancy falls under \textbf{name inconsistency}.
\end{minipage}  \\ \midrule
\begin{minipage}[t]{0.17\linewidth}\centering \textbf{C2.} Type\\Inconsistency\end{minipage} & 
\begin{minipage}[t]{0.5\linewidth}
def scan(self, package=None, ... , **kw,):\\
\hspace*{1em}package = self.maybe\_dotted(package) \\
\hspace*{1em}if package is None: \\
\hspace*{2em}package = \textcolor{red}{return\_package()} \\
\hspace*{1em}ctorkw = {'config': self} \\
\hspace*{1em}ctorkw.update(kw) \\
\hspace*{1em}\textcolor{teal}{// Some code $\dotsc$} \\ 
\end{minipage}  &
\begin{minipage}[t]{0.25\linewidth}
[...] The `return\_package' function returns the name of the calling module’s package as a \textcolor{red}{string}. [...]
\end{minipage} & \begin{minipage}[t]{0.4\linewidth}
When examining the actual functionality of `return\_package', we find that it actually \textbf{returns the calling package as a `sys.modules' object}. This discrepancy falls under \textbf{type inconsistency}.
\end{minipage}  \\ \midrule
\begin{minipage}[t]{0.17\linewidth}\centering \textbf{C3.} Functionality\\Inconsistency\end{minipage} & 
\begin{minipage}[t]{0.5\linewidth}
def \_is\_asgi\_app(app): \\
\hspace*{1em}app\_args = \textcolor{red}{inspect.getfullargspec(app)} \\
\hspace*{1em}num\_app\_args = len(app\_args) \\
\hspace*{1em}\textcolor{teal}{// Some code $\dotsc$} \\
\hspace*{1em}if app\_args[0] in {'cls', 'self'}: \\
\hspace*{2em}num\_app\_args -= 1 \\
\hspace*{1em}is\_asgi = num\_app\_args == 3 \\
\hspace*{1em}return is\_asgi
\end{minipage} &
\begin{minipage}[t]{0.25\linewidth}
[...] It uses the `inspect' module \textcolor{red}{to retrieve the full argument specification of the function or method passed as `app'.} [...]
\end{minipage} & \begin{minipage}[t]{0.4\linewidth}
This summary insufficiently and inaccurately describes the dependency function's behavior. The actual functionality of \textbf{`inspect.getfullargspec()' does not merely retrieve the full arguments.} This function initializes the app parameter and then returns the full set of argument values.
\end{minipage}  \\ \midrule
\begin{minipage}[t]{0.17\linewidth}\centering \textbf{C4.} Context\\Irrelevant\end{minipage} & 
\begin{minipage}[t]{0.5\linewidth}
\hspace*{1em}\textcolor{teal}{// Some code $\dotsc$} \\
\hspace*{1em}if settings.USE\_TZ and is\_naive(dt): \\
\hspace*{2em}return make\_aware(dt, \textcolor{red}{timezone=timezone.utc}) \\
\hspace*{1em}\textcolor{teal}{// Some code $\dotsc$} \\
\hspace*{1em}return dt \end{minipage}
&
\begin{minipage}[t]{0.25\linewidth}
[...] \textcolor{red}{UTC (Coordinated Universal Time) timezone is the primary time standard by which the world regulates clocks and time.}
\end{minipage} & \begin{minipage}[t]{0.4\linewidth}
The generated summary \textbf{includes an additional explanation about the UTC timezone, which is unrelated to the function’s actual behavior}. This constitutes irrelevant context in the summary.
\end{minipage}  \\ \bottomrule
\end{tabular}
\end{adjustbox}
\caption{Examples and explanations about four major patterns of factual inconsistency detected by human annotators. Factual inconsistencies between the input code and the generated summary are highlighted in \textcolor{red}{red}.}
\label{tab:cases_example}
\end{table*}

\subsection{Code-Related Information Searching} \label{A.2}
This section provides a detailed description of the code-related information searching method introduced in Section~\ref{3.2}.


\paragraph{Step1: Project context graph construction}
When constructing a project context graph, we build a heterogeneous directed acyclic graph (DAG) of code entities and their dependency relationships based on abstract syntax trees (AST)~\citep{8812062} and static program analysis techniques.
Each node stores metadata such as the entity’s AST type, code context, and docstring. The edges represent the flow of variables, indicating where they originate and where they are propagated. To capture variable relationships in a more structured way, we apply five relation types to the edges, as defined by \citet{cheng-etal-2024-dataflow}:

\begin{itemize}
  \item \textbf{Assign} relation is a one-to-one correspondence in an assignment statement, which controls variable creation and mutation
  \item \textbf{As} relation is from with or except statements and similar to the \textit{assigns} relation.
  \item \textbf{Refers} relation that represents a reference to an existing variable or its attribute.
  \item \textbf{Typeof} relation indicates the data type of the (return) value of a variable or function.
  \item \textbf{Inherits} relation is an implicit data dependency relation since a subclass inherits all the class members of its base classes.
\end{itemize}

Our DFG is a heterogeneous directed acyclic graph $G=\{(h,r,t)|h,t \in E, r \in R\}$, where $E$ denotes the entity set, $R$ denotes the type-sensitive relation set, and the triplet $(h,r,t)$ represents the head entity $h$ pointing to the tail entity $t$ with the relation $r$.

\paragraph{Step2: Code-related information selection}
In the next step, we extract relevant node information from the project context graph based on the input code's dependency relationships.

In this paper, information related to each entity in the input code is selected through the following steps for efficiency.
\textbf{(1)} We perform a depth-first search (DFS) on the project context graph, starting from each entity, to traverse all related nodes connected via dependency edges. 
\textbf{(2)} If any of the related nodes for a given entity are \texttt{import\_statement} (\texttt{import\_declaration for Java language}, they are transformed into the form \texttt{(module, name)} or \texttt{(module, name.attr)}. 
For example,
\texttt{``from utils.collection\_utils import key\_prefix''} is converted to \texttt{``(utils.collection\_utils, key\_prefix)''}.
\textbf{(3)} If the \texttt{module} path exists within the project directory, the given entity is considered to have a \textbf{cross-file dependency}. In this case, we perform an exact match search that precisely corresponds to the \texttt{name} or \texttt{name.attr} within that \texttt{module} file. 
If the searched corresponding node type is \texttt{class\_definition}, \texttt{function\_definition}, or \texttt{assignment}, we select the corresponding node context (\texttt{class\_declaration}, \texttt{method\_declaration}, or \texttt{field\_declaration} for Java language.).
\textbf{(4)} If the \texttt{module} path does not exist within the project directory and is imported from external sources such as Python built-in or third-party libraries, the given entity is considered to have an \textbf{external dependency}. In such cases, we select the corresponding description for the given \textit{name} or \textit{name.attr} from a predefined external API documentation, collected from ODEX~\citep{wang2022execution}, DS-1000~\citep{lai2023ds}, and the official Python and Java documentation site\footnote{Python : \url{https://docs.python.org/3/}, Java: \url{https://docs.oracle.com/en/java/}}.
\textbf{(5)} Finally, if none of the related nodes are of the \texttt{import\_statement} type, it indicates that the dependency information for the given node is located within the same file as the input code (\textbf{internal-file dependency}). In this case, if any related nodes are of type \texttt{class\_definition}, \texttt{function\_definition}, or \texttt{assignment} (\texttt{class\_declaration}, \texttt{method\_declaration}, or \texttt{field\_declaration} for Java language.), we select the corresponding node context.
\textbf{(6)} The information obtained through steps (1–5) is prepended to the input code in the format ``\# \{name\} \# \{content\}''. If the information corresponds to a class, the content contains the class docstring. For functions or variables, the content includes the code body.

\subsection{Segment-Level Evaluator} \label{A.3}
At the sentence-level factual consistency evaluation stage, \textbf{we focus on defining sentence-level judgments in terms of an objectively verifiable ``presence or absence of error''}, designing a clear binary prompt in which a sentence is assigned a score of 0 if any factual inconsistency is detected with respect to a given criterion.

The prompt for segment-level factual inconsistency criterion evaluation is described below. The \textcolor{blue}{\{criterion\}} and \textcolor{blue}{\{explanation\}} fields contain the name (e.g., Name Inconsistency) and definition described in Table~\ref{tab:criteria}. This prompt is input into the model, which outputs either 1 or 0 at segment and criteria level.

\begin{tcolorbox}[
  title=Segment-level criterion evaluation prompt,
  colback=gray!15,
  colframe=gray!60,
  coltitle=black,
  fonttitle=\bfseries,
  boxrule=0.1pt,
  boxsep= 2pt,
  left=3pt,
  right=3pt,
  top=2pt,
  bottom=2pt,
  breakable
]
\textbf{[System Prompt]} \\
You will be given one summary text written for a source code. Your task is to evaluate the summary from \textcolor{blue}{\{criterion\}} aspect.
Please make sure you read and understand these instructions carefully. Please keep this document open while reviewing, and refer to it as needed. \\

Evaluation Criteria: \\
\textcolor{blue}{\{criterion\}} (1 or 0) -- \textcolor{blue}{\{explanation\}} \\

Evaluation Steps: \\
1. Read the CODE carefully and understand its main intent. \\
2. Read the code summary text and check if it accurately describes the code. \\
3. Evaluate whether \textcolor{blue}{\{criterion\}} exists, where ``1'' means ``\textcolor{blue}{\{criterion\}} does not exist'' and ``0'' means ``\textcolor{blue}{\{criterion\}} exists'' based on the Evaluation Criteria.
\\ \\
\textbf{[User Prompt]} \\
\#\# CODE: \\
(Related Information) \textcolor{blue}{\{related\_information\}} \\
(Input Code) \textcolor{blue}{\{input\_code\}} \\
\#\# SUMMARY TEXT: \textcolor{blue}{\{segment\}} \\
\#\# SCORE (score only):
\end{tcolorbox}

\section{Additional Details About Experimental Setups} \label{B}

\subsection{More Details About Benchmark Construction Process} \label{B.1}

\paragraph{Summary Generation Prompt:}
The goal of our benchmark is \textit{(1)} to evaluate whether ReFEree can distinguish the four factual consistency criteria (C1-C4) in a manner similar to human judgement, and \textit{(2)} to assess whether ReFEree's final scores correlate well with human ratings. 

To properly evaluate these capabilities, the benchmark must contain a sufficiently balanced number of gold labels across all four criteria. However, when using only naturally generated summaries as-is, we observe two issues (as shown in the Table~\ref{tab:empirical_analysis_v2}; The overall frequency of factual inconsistencies is relatively low, and there exists a significant label imbalance across the four criteria. This imbalance makes it difficult to reliably measure criterion-level evaluation ability. Therefore, to ensure sufficient coverage for all evaluation dimensions, we intentionally incorporated inconsistency patterns into the prompt when generating hallucinated summaries.

Importantly, these inconsistency patterns are not arbitrarily designed. In Section~\ref{3.1}, we analyzed naturally generated summaries from three LLMs and identified four representative, naturally occurring error patterns. These empirically observed patterns were then reflected in the construction of hallucinated summaries. In other words, the inconsistency types included in our benchmark are grounded in real-world error distributions rather than being artificially invented rubric-specific artifacts.

Below is the full prompt for generating code summaries with intentional factual inconsistencies. We use the ChatGPT model (\texttt{chatgpt-4o-latest}), with the following hyperparameter settings: temperature = \texttt{0.6}, top-p = \texttt{0.9}, and max new tokens = \texttt{256}. The total cost of generating hallucinated summaries for 2,075 samples amounts to approximately \$250. \textbf{\texttt{\{lang\}}} corresponds to either Python or Java.

\begin{tcolorbox}[
  title=Hallucinated summary generation prompt,
  colback=gray!15,
  colframe=gray!60,
  coltitle=black,
  fonttitle=\bfseries,
  boxrule=0.1pt,
  boxsep= 2pt,
  left=3pt,
  right=3pt,
  top=2pt,
  bottom=2pt,
  breakable  
]
\textbf{[System Prompt]} \\
You are an expert in understanding \texttt{\textbf{\{lang\}}} code. As a \texttt{\textbf{\{lang\}}} expert, please generate a detailed and informative hallucinated summary of the following \texttt{\textbf{\{lang\}}} code. The hallucinated summary contains context that sounds plausible but does not accurately reflect the actual implementation or behavior of the code. \\
You SHOULD generate a summary that includes at least one (one or more) hallucinated sentence corresponding explicitly to one of the following five factual inconsistency cases, respectively.:
\bgminipage{orange}{1. The name of a function, class, or variable mentioned in the text does not match the actual identifier used in the code.} \\
\bgminipage{green}{$<$Demonstration$>$\\
\#\# code: \texttt{\textbf{\{code\}}} \\
\#\# Hallucinated Summary: \texttt{\textbf{\{summary\}}}} \\
\bgminipage{orange}{2. The described return type or variable type in the text is inconsistent with the actual type used or inferred in the code.}\\
\bgminipage{green}{$<$Demonstration$>$} \\
\bgminipage{orange}{3. The described functionality or purpose of the code in the text does not accurately reflect what the Python code actually implements.} \\
\bgminipage{green}{$<$Demonstration$>$} \\
\bgminipage{orange}{4. The described text contains content that is unnecessary or unrelated to the input code.}\\
\bgminipage{green}{$<$Demonstration$>$} \\
You should try your best to make the hallucinated summary. Provide ONLY the summary texts. Do not include any other codes or notes.
\\ \\
\textbf{[User Prompt]} \\
\#\# CODE: \\
(Related Information) \textcolor{blue}{\{related\_information\}} \\
(Input Code) \textcolor{blue}{\{input\_code\}} \\
\#\#\# HALLUCINATED SUMMARY:
\end{tcolorbox}

\paragraph{Factual Consistency Labeling Process:}
As many prior studies~\citep{7298824,desmond2021increasing, zhang2023peanut, kim2024meganno+} have noted, the primary advantage of human–AI collaborative labeling systems is that AI and humans can complement each other’s weaknesses, thereby improving annotation quality.
Human annotators use their expertise to verify and correct hallucinations and factual errors that may occur in LLM-generated labels, while LLM-based labelers rapidly propose initial candidate labels for large-scale data, reducing the excessive annotation time and subjectivity that arise when relying solely on humans.

\textbf{First, three LLMs with different roles determine the candidate label through majority voting.} If all three LLMs assign different scores during summary-level labeling (since there are five possible labels), the score closest to the mean is selected as the candidate label. The two greyboxes below show the full prompts used to determine candidate labels for summary-level and segment-level annotations, respectively.
We use ChatGPT (\texttt{chatgpt-4o-latest}) with the following hyperparameter settings: temperature = \texttt{0.1}, top-p = \texttt{0.9}, and max new tokens = \texttt{16}. The total cost of predicting labels for 2,055 samples amounts to approximately \$300 at the summary level and \$800 at the segment level. \textbf{\texttt{\{lang\}}} corresponds to either Python or Java and each LLM is assigned a distinct \textcolor{blue}{\{role\}}: Code Editor, Code Reviewer, or Original Code Author.

\begin{tcolorbox}[
  title=Summary-level labeling prompt,
  colback=gray!15,
  colframe=gray!60,
  coltitle=black,
  fonttitle=\bfseries,
  boxrule=0.1pt,
  boxsep= 2pt,
  left=3pt,
  right=3pt,
  top=2pt,
  bottom=2pt,
  breakable
]
\textbf{[System Prompt]}\\
As a \textcolor{blue}{\{role\}}, your task is to rate the factual consistency of a summarized text generated from a \texttt{\textbf{\{lang\}}} code and related information. \\ 
Factual Consistency: Guarantee that the summary remains consistent with the CODE and RELATED INFORMATION, accurately capturing its primary functionality and logic without adding any unrelated content. \\ 
Please refer to the CODE, RELATED INFORMATION, and the SUMMARY, and then assign a factual consistency score based on the following grading rubric. \\

\# SCORE RUBRIC: \\
- 5: Highly Consistent : All information in the generated content can be verified in the CODE and RELATED INFORMATION. \\
- 4: Very Consistent : Most information in the generated content can be verified in the CODE and RELATED INFORMATION, with one minor item that wouldn’t negatively impact the reader’s understanding. \\
- 3: Moderately Consistent : More than one piece of information in the generated content cannot be verified in the CODE and RELATED INFORMATION, but none of these inaccuracies would negatively impact the reader’s understanding. \\
- 2: Somewhat Inconsistent : One or more pieces of information in the generated content are factually inaccurate and cannot be verified in the CODE and RELATED INFORMATION, some, or all of which would negatively impact the reader’s understanding. \\
- 1: Highly Inconsistent : Most or all of the information in the generated content is inaccurate, cannot be verified in the CODE and RELATED INFORMATION, and would negatively impact the reader’s understanding. \\

Generate ONLY the score. Do not include any other codes, or notes.
\\ \\
\textbf{[User Prompt]} \\
\#\# CODE: \\
(Related Information) \textcolor{blue}{\{related\_information\}} \\
(Input Code) \textcolor{blue}{\{input\_code\}} \\
\#\# SUMMARY: \textcolor{blue}{\{summary\}} \\
\#\# SCORE:
\end{tcolorbox}

\begin{tcolorbox}[
  title=Segment-level labeling prompt,
  colback=gray!15,
  colframe=gray!60,
  coltitle=black,
  fonttitle=\bfseries,
  boxrule=0.1pt,
  boxsep= 2pt,
  left=3pt,
  right=3pt,
  top=2pt,
  bottom=2pt,
  breakable
]
As a \textcolor{blue}{\{role\}}, you ensure the factual consistency with the \texttt{\textbf{\{lang\}}} code and the given sentence, which is a part of the code summary. \\
The given text should be precise and only include verifiable information that is explicitly stated in the source code, so do not make any assumptions or derive any thoughts. \\

Given the CODE, RELATED INFORMATION, and the SUMMARY TEXT, which is the part of the summary, your objective is to evaluate whether the sentence contains \textcolor{blue}{\{criterion\}} that \textcolor{blue}{\{explanation\}}.
If the sentence strictly corresponds to a \textcolor{blue}{\{criterion\}}, you should output 0; otherwise, output 1.
\\ \\
\textbf{[User Prompt]} \\
\#\# CODE: \\
(Related Information) \textcolor{blue}{\{related\_information\}} \\
(Input Code) \textcolor{blue}{\{input\_code\}} \\
\#\#\# SUMMARY TEXT: \textcolor{blue}{\{segment\}} \\
\#\#\# SCORE:
\end{tcolorbox}

\begin{figure*}[t]
    \centering
    \begin{subfigure}{\linewidth}
        \centering
        \includegraphics[width=1\linewidth]{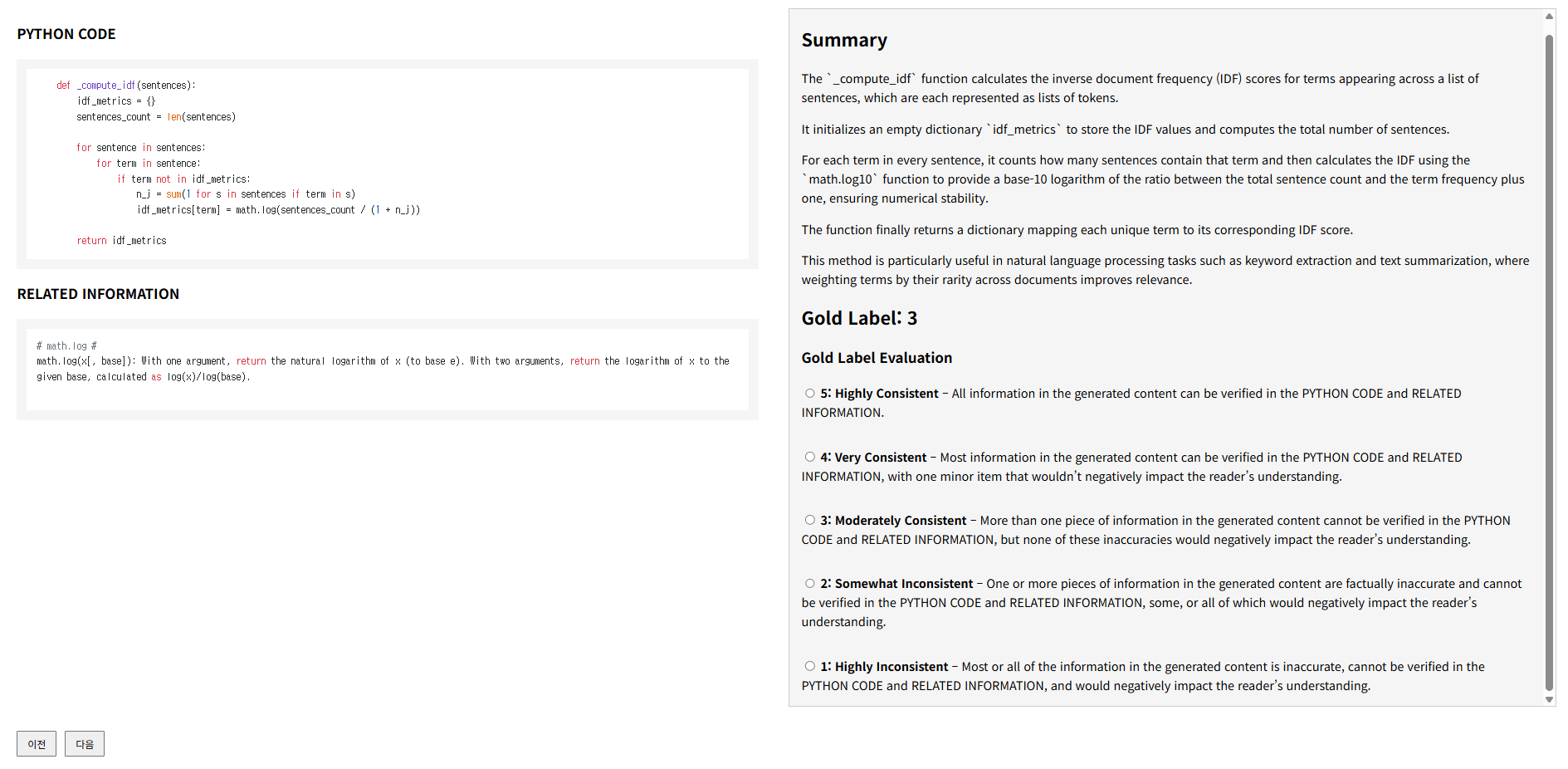}
        \caption{Summary-level human annotation interface}
        \label{fig:human_eval_img_a}
    \end{subfigure}
    \\[1em]
    \begin{subfigure}{\linewidth}
        \centering
        \includegraphics[width=1\linewidth]{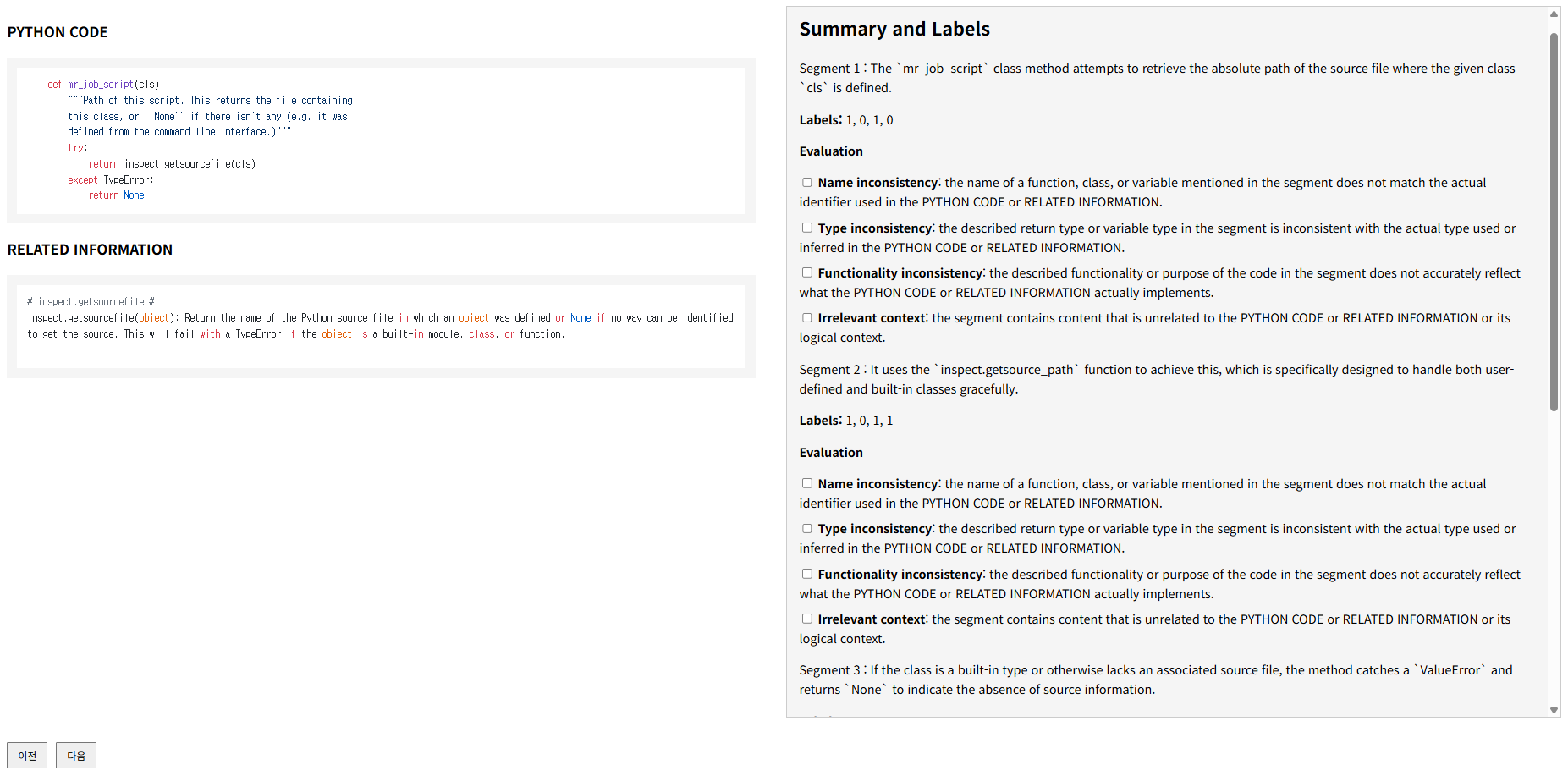}
        \caption{Segment-level human annotation interface}
        \label{fig:human_eval_img_b}
    \end{subfigure}
    \caption{\textbf{Human evaluation interfaces:} The \textbf{upper} image shows the summary-level human annotation interface, while the \textbf{lower} image shows the segment-level interface.}
    \label{figure:human_eval_img}
\end{figure*}

\textbf{Then, three human annotators subsequently post-edit the candidate label to determine the final label.} They are instructed to retain the label if they agree with the candidate label, or to revise it if they do not. The final label is determined according to the following three rules:
(1) If at least two annotators agree with the candidate label, the label is retained.
(2) If at least two annotators revise to the same label, the label is edited accordingly.
(3) For summary-level labeling: If all annotators assign different labels, the final label is determined through further discussion among the three annotators.
Figure~\ref{figure:human_eval_img} shows the platform used for the human annotation process. 


\paragraph{Dataset Statistics and Examples: }
Table~\ref{tab:statistics} presents the label distribution of our benchmark. The benchmark consists of 2,055 code summaries (1,825 Python and 230 Java) annotated at both summary and segment levels. At the summary level, each sample is labeled on a 5-point scale ranging from 1 (highly inconsistent) to 5 (highly consistent). At the segment level, the benchmark contains 11,804 annotations (10,415 Python and 1,389 Java), where each sentence is labeled as CORRESPOND (1) or NOT CORRESPOND (0) for each of the four factual inconsistency criteria (C1–C4). Table~\ref{tab:benchmark_example} provides some examples of our benchmark.

\begin{table}[t]
\begin{adjustbox}{max width=\linewidth}
\begin{tabular}{ccccc}
\toprule
\multicolumn{5}{c}{\cellcolor[HTML]{EDEDED}{\textbf{Summary-Level}}}                         \\ \midrule
\multicolumn{1}{c|}{\textbf{Label}}      & \multicolumn{2}{c|}{\textbf{Python}}         & \multicolumn{2}{c}{\textbf{Java}}  \\ \midrule
\multicolumn{1}{l|}{1 (Highly Inconsistent)}     & \multicolumn{2}{c|}{16}                      & \multicolumn{2}{c}{7}              \\
\multicolumn{1}{l|}{2 (Somewhat Inconsistent)}   & \multicolumn{2}{c|}{829}                     & \multicolumn{2}{c}{54}             \\
\multicolumn{1}{l|}{3 (Moderately Consistent)} & \multicolumn{2}{c|}{435}                     & \multicolumn{2}{c}{74}             \\
\multicolumn{1}{l|}{4 (Very Consistent)}       & \multicolumn{2}{c|}{488}                     & \multicolumn{2}{c}{87}             \\
\multicolumn{1}{l|}{5 (Highly Consistent)}     & \multicolumn{2}{c|}{57}                      & \multicolumn{2}{c}{8}              \\ \midrule
\multicolumn{1}{c|}{\textbf{Total}}      & \multicolumn{2}{c|}{\textbf{1,825}}          & \multicolumn{2}{c}{\textbf{230}}   \\ \midrule
\multicolumn{5}{c}{\cellcolor[HTML]{EDEDED}{\textbf{Segment-Level}}}                                                         \\ \midrule
\multicolumn{1}{c|}{\textbf{Label}}      & \textbf{1} & \multicolumn{1}{c|}{\textbf{0}} & \textbf{1}       & \textbf{0}      \\ \midrule
\multicolumn{1}{l|}{C1 (Name Inconsistent)}                  & 3,815      & \multicolumn{1}{c|}{6,600}      & 983              & 406             \\
\multicolumn{1}{l|}{C2 (Type Inconsistent)}                  & 2,844      & \multicolumn{1}{c|}{7,571}      & 1,131            & 258             \\
\multicolumn{1}{l|}{C3 (Functionality Inconsistent)}                  & 3,975      & \multicolumn{1}{c|}{6,440}      & 700              & 689             \\
\multicolumn{1}{l|}{C4 (Context Irrelevant}                  & 4,690      & \multicolumn{1}{c|}{5,725}      & 635              & 754             \\ \midrule
\multicolumn{1}{c|}{\textbf{Total}}      & \multicolumn{2}{c|}{\textbf{10,415}}         & \multicolumn{2}{c}{\textbf{1,389}} \\ \bottomrule
\end{tabular}
\end{adjustbox}
\caption{Label distribution of our benchmark. Summary-level labels range from 1 (highly inconsistent) to 5 (highly consistent). Segment-level labels indicate whether each criterion (C1–C4) is CORRESPOND (1) or NOT CORRESPOND (0).}
\label{tab:statistics}
\end{table}

\paragraph{Regarding Potential Risks of Systemic Bias During Annotation:}
To mitigate potential annotator bias from LLM-provided candidate scores, our labeling pipeline treats model scores as non-binding references rather than ground truth. Annotators are explicitly instructed that scores may be freely overridden, and final labels are determined by majority voting across three LLMs and three human annotators, ensuring that neither human nor model judgment dominates.

To empirically validate this design, we compare labels produced under our \textit{Human-AI} collaboration system against labels collected entirely by humans from scratch (\textit{human-only}). The two settings yield an agreement rate of 89.5\%, and among the 10.5\% of mismatched cases, the mean absolute score difference is only 0.105 - with all disagreements falling within $\pm$1 point and no case exceeding a 2-point gap. Label correlation is consistently high (Pearson = 0.9401, Spearman = 0.9499, Kendall = 0.9242), confirming that the \textit{Human-AI} approach preserves annotation quality while substantially reducing cost and time.

\subsection{Details About Baselines} \label{B.2} 
We select 13 commonly used evaluation methods, 8 reference-based and 5 reference-free methods, from code summarization tasks to compare with our proposed method.

\paragraph{1. Reference-based methods:} We use the English descriptions as reference summaries.

\noindent\textbf{ROUGE~\citep{lin-2004-rouge}} ROUGE measures the overlap of n-grams between the generated output and reference summaries. In this paper, we use ROUGE-1/2/L f1 score for baselines.

\noindent\textbf{BLEU~\citep{papineni2002bleu}} measures the n-gram precision between the generated text and references with an additional brevity penalty to discourage short outputs. 

\noindent\textbf{BERTScore~\citep{zhang2019bertscore}} BERTScore computes similarity between generated and reference texts using contextual embeddings from pre-trained BERT models. It captures semantic similarity better than traditional lexical metrics.

\noindent\textbf{METEOR~\citep{banerjee-lavie-2005-meteor}} METEOR, a recall-oriented metric, evaluates how well the model captures reference content by matching words between candidate and reference sentences and computing the harmonic mean of precision and recall. 

\noindent\textbf{Sentencebert~\citep{reimers2019sentence}} SentenceBERT encodes sentences into dense vector representations to compute semantic similarity via cosine distance or Euclidean distance.

\paragraph{2. Reference-free methods:} We evaluate the overall factual consistency between the input code and the entire summary with 5 baselines using the same LLM, GPT-4.1-mini (\texttt{gpt-4.1-mini-2025-04-14}), and the hyperparameters are set as follows: temperature = 0.1, top-p = 0.9, top-k = 50, and max new tokens = 4. 
Since LLM-Judge, G-Eval, and FactScore are originally designed as evaluation methods for the NLP domain, we modify their prompts to ensure applicability to the code domain.

\noindent\textbf{SIDE~\citep{mastropaolo2024evaluating}} SIDE evaluates the semantic fitness of code summaries by using contrastive learning to distinguish between characteristics of suitable and unsuitable summaries for a given code. We used the pre-trained SIDE model from GitHub~\footnote{\url{https://github.com/antonio-mastropaolo/code-summarization-metric}}.

\noindent\textbf{LLM-Judge~\citep{zheng2023judging}} This paper was the first to propose the LLM-as-a-Judge framework. We utilize the default prompt for single-answer grading from this paper, inputting the code and summary to evaluate consistency. The prompt is shown below.

\begin{tcolorbox}[
  title=LLM-judge prompt,
  colback=gray!15,
  colframe=gray!60,
  coltitle=black,
  fonttitle=\bfseries,
  boxrule=0.1pt,
  boxsep= 2pt,
  left=3pt,
  right=3pt,
  top=2pt,
  bottom=2pt,
  breakable
]
\textbf{[System Prompt]}\\
Please act as an impartial judge and evaluate the quality of the response provided by an AI assistant to the user question displayed below.  \\
Your evaluation should consider the factually consistency that the summary remains consistent with the original code, accurately capturing its primary functionality and logic without adding any unrelated content.\\
Please rate the response on a scale of 1 to 5.
\\ \\
\textbf{[User Prompt]} \\
\#\# CODE: \textcolor{blue}{\{input\_code\}} \\
\#\# SUMMARY TEXT: \textcolor{blue}{\{summary\}} \\
\#\# SCORE (score only):
\end{tcolorbox}

\noindent\textbf{G-Eval~\citep{liu-etal-2023-g}} This method is a framework that utilizes Large Language Models (LLMs) with Chain-of-Thought (CoT) and a form-filling paradigm to assess the quality of Natural Language Generation (NLG) outputs. We adapt the prompt to be suitable for code summary evaluation, as shown below.

\begin{tcolorbox}[
  title=G-eval prompt,
  colback=gray!15,
  colframe=gray!60,
  coltitle=black,
  fonttitle=\bfseries,
  boxrule=0.1pt,
  boxsep= 2pt,
  left=3pt,
  right=3pt,
  top=2pt,
  bottom=2pt,
  breakable
]
\textbf{[System Prompt]}\\
You will be given one summary written for a source code. Your task is to rate the summary from factually consistency aspect.
Please make sure you read and understand these instructions carefully. Please keep this document open while reviewing, and refer to it as needed.

Evaluation Criteria:\\
Consistency (1-5) -  Guarantee that the summary remains consistent with the original code, accurately capturing its primary functionality and logic without adding any unrelated content.\\

Evaluation Steps:\\
1. Read the CODE carefully and understand its main intent.\\
2. Read the code summary and check if it accurately describe the code.\\
3. Assign a score for factually consistency on a scale of 1 to 5, where 1 is the lowest and 5 is the highest based on the Evaluation Criteria.
\\ \\
\textbf{[User Prompt]} \\
\#\# CODE: \textcolor{blue}{\{input\_code\}} \\
\#\# SUMMARY TEXT: \textcolor{blue}{\{summary\}} \\
\#\# SCORE (score only):
\end{tcolorbox}

\noindent\textbf{Factscore~\citep{min-etal-2023-factscore}} This method is a fine-grained method proposed in the NLP fields that breaks a generation into a series of atomic facts and computes the percentage of atomic facts supported by a reliable knowledge source. For use in code summary evaluation, we break down the code summary into a series of atomic facts and compute the percentage of consistency with the code. We first generate these atomic facts using the prompt: \texttt{``Please breakdown the following sentence into independent facts.''} Subsequently, each atomic fact is evaluated using the prompt: \texttt{``Answer the question based on the given context. Context: \{code\} Input: \{summary\} True or False? Output:''}. The input code is provided as the knowledge source instead of general background knowledge.

\noindent\textbf{CODERPE~\citep{wu2024can}} This method uses a Role-Player system prompt designed to quantify the quality
of generated code summaries. We prompt an LLM agent to
perform a role, code reviewer. This task involves evaluating the quality of generated code summaries, specifically along the dimension of consistency. The prompt is shown below.

\begin{tcolorbox}[
  title=CODERPE prompt,
  colback=gray!15,
  colframe=gray!60,
  coltitle=black,
  fonttitle=\bfseries,
  boxrule=0.1pt,
  boxsep= 2pt,
  left=3pt,
  right=3pt,
  top=2pt,
  bottom=2pt,
  breakable
]
\textbf{[System Prompt]}\\
As a code reviewer, you ensure the factually consistency of the code summary. \\

You will be given one summary written for a source code. Your task is to rate the summary from factually consistency aspect.
Please make sure you read and understand these instructions carefully. Please keep this document open while reviewing, and refer to it as needed. \\

Evaluation Criteria: \\
Consistency (1-5) -  Guarantee that the summary remains consistent with the original code, accurately capturing its primary functionality and logic without adding any unrelated content. \\

Evaluation Steps: \\
1. Read the CODE carefully and understand its main intent. \\
2. Read the code summary and check if it accurately describe the code. \\
3. Assign a score for factually consistency on a scale of 1 to 5, where 1 is the lowest and 5 is the highest based on the Evaluation Criteria.
\\ \\
\textbf{[User Prompt]} \\
\#\# CODE: \textcolor{blue}{\{input\_code\}} \\
\#\# SUMMARY TEXT: \textcolor{blue}{\{summary\}} \\
\#\# SCORE (score only):
\end{tcolorbox}

\subsection{Implementation Details} \label{B.3}
Our method supports various LLMs, including both closed-source and open-source models, as segment-level criterion evaluators. In our main experiments, we use OpenAI's GPT-4.1-mini to ensure a fair comparison with existing baselines under the same setting. All evaluations are conducted with consistent hyperparameters: temperature = \texttt{0.1}, top-p = \texttt{0.9}, top-k = \texttt{50}, and max new tokens = \texttt{4}.

\section{Additional Experimental Results} \label{C}

\subsection{Information Searching Ablations}  \label{5.6}

We additionally presented experimental results comparing evaluations using 0, 1, and 2-hop context in Python benchmarks. As shown in Table~\ref{tab:hop}, the 1-hop setting achieves the highest average correlation (0.459), outperforming both 0-hop (0.404) and 2-hop (0.448). While using 2-hop context did improve correlation compared to 0-hop, it still underperformed the 1-hop setting. The key insights derived from this analysis are as follows:

\noindent 1) Our analysis shows that most factual inconsistencies can be correctly determined based on information from directly invoked entities (depth-1).  
\noindent 2) When expanding retrieval to 2-hop or deeper, the additional information such as transitive dependencies, internal implementation details or indirect call-chain information typically includes. However, our experimental results show that this additional information is rarely mentioned in actual code summaries and is not typically considered in human factuality judgments. Instead, it tends to introduce irrelevant context, which can make the evaluation less stable and less accurate.


\begin{table}[t]
\begin{adjustbox}{max width=\linewidth}
\begin{tabular}{lcccc}
\toprule
\textbf{Context setting}   & $r_p$ & $r_s$ & $\tau$ & \textbf{Average} \\ \midrule
0-hop (w/o info) & 0.432                                & 0.432                                 & 0.349                                & 0.404                                \\
\textbf{1-hop (ours)}      & \textbf{0.497}                       & \textbf{0.489}                        & \textbf{0.390}                       & \textbf{0.459}                       \\
2-hop                      & 0.491                                & 0.474                                 & 0.377                                & 0.448                                \\ \bottomrule
\end{tabular}
\end{adjustbox}
\caption{Comparison of correlation coefficients across different search depths for code-related information.}
\label{tab:hop}
\end{table}

\subsection{Time and computational costs} \label{C.4}
We compare the per-sample execution time and cost between baseline methods and our approach, with results presented in Table~\ref{tab:time_cost}. Most reference-based methods incur no API cost, making them computationally efficient. However, their low correlation with human judgment makes them unsuitable for evaluating long, descriptive project-level code summaries.
Our method, ReFEree, requires longer execution time per sample compared to simpler baselines due to its fine-grained evaluation process. However, at the evaluation stage, ensuring reliability and accuracy is more critical than marginal gains in efficiency. For instance, G-Eval is fast (approximately 1.11 seconds per sample) but shows limited correlation with human judgment. FactScore adopts a similar fine-grained framework and requires approximately 9.18 seconds per sample, yet achieves lower correlation (0.391 in Python, 0.355 in Java). In contrast, ReFEree achieves the highest correlation (0.459 in Python, 0.480 in Java) within a comparable time range, demonstrating a favorable trade-off between computational cost and evaluation quality.

\begin{table}[t]
\centering
\begin{adjustbox}{max width=\linewidth}
\begin{tabular}{l|cc}
\toprule
\textbf{Methods}     & \textbf{Time/s} & \textbf{Cost/s}   \\ \midrule
ROUGE-1          & 0.00   & 0.0000 \\
ROUGE-2          & 0.00   & 0.0000 \\
ROUGE-L          & 0.00   & 0.0000 \\
BLEU             & 1.12   & 0.0000 \\
METEOR           & 0.02   & 0.0000 \\
BERTScore        & 0.03   & 0.0000 \\
SBCS             & 0.02   & 0.0000 \\
SBED             & 0.03   & 0.0000 \\
SIDE             & 0.08   & 0.0000 \\ \midrule
LLM-judge        & 0.34   & 0.0002 \\
G-Eval           & 1.11   & 0.0002 \\
Factscore        & 9.18   & 0.0058 \\
CODERPE          & 0.53   & 0.0002 \\
\textbf{ReFEree} & 10.24  & 0.0042 \\ \bottomrule
\end{tabular}
\end{adjustbox}
\caption{Comparison of inference time (seconds) and cost (\$) per sample across evaluation methods. Average time and cost are computed over 2,055 code summary evaluations in our benchmark.}
\label{tab:time_cost}
\end{table}

\subsection{Examples of \textit{ReFEree}} \label{C.5}
The results of evaluating factual consistency through our method are shown in Tables~\ref{tab:cases_example_1} and Table~\ref{tab:cases_example_2}. For both examples, existing methods such as ROUGE-L, BERTScore, and SIDE produce results that differ significantly from human labels in their evaluation of the factual consistency between the code and the generated summary. However, the \textit{ReFEree} method outputs consistency scores that are closer to human label evaluations. Additionally, our method can explain which parts of the generated summary and what types of inconsistencies occur.

\subsection{Applicable to Human-Written Summaries} \label{C.3}
Our method focuses on accurately evaluating factual inconsistencies in LLM-generated code summaries. However, ReFEree is not limited to LLM outputs. Our method can also evaluate summaries written by real developers. 
We additionally collected 183 human-written docstrings from the Deveval dataset along with their corresponding human-annotated factual consistency scores. Based on this data, we performed a comparative analysis against existing baselines. 

As shown in Table~\ref{tab:human_eval}, human-written summaries rarely contain factual errors. Consequently, the human scores are highly skewed. This results in extremely low label variance, making meaningful correlation difficult to obtain. Nevertheless, ReFEree still achieved higher correlation than the baseline methods. This result suggests that ReFEree can meaningfully evaluate real human-written documentation and quantitatively capture the intuitive observation that high-quality human summaries tend to exhibit greater factual consistency.

\begin{table}[t]
\centering
\begin{adjustbox}{max width=\linewidth}
\begin{tabular}{l|c|ccc|c}
\toprule
\textbf{Method} & \textbf{Score} & $r_p$        & $r_s$   & $\tau$ & \textbf{Average} \\ \midrule
Human (1–5) & 4.938 & x & x & x & x \\
ROUGE-L (0–1) & 0.208 & 0.006 & 0.047 & 0.039 & 0.030 \\
BLEU (0–1) & 0.030 & 0.062 & 0.061 & 0.054 & 0.059 \\
G-Eval (1–5) & 3.043 & 0.013 & 0.003 & 0.003 & 0.006 \\ \midrule
ReFEree (0–1) & \textbf{0.938} & \textbf{0.326} & \textbf{0.298} & \textbf{0.287} & \textbf{0.304} \\
 \bottomrule
\end{tabular}
\end{adjustbox}
\caption{Evaluation results on 183 human-written docstrings from the DevEval dataset.}
\label{tab:human_eval}
\end{table}

\subsection{Criterion Weight Configuration} \label{C.2}
In our scoring aggregation, we assign equal weights to each criterion (C1–C4) by default. This design choice does not imply that all criteria are equally important, but rather reflects a configuration that treats all criteria evenly without introducing additional hyperparameters. To explore whether weighted aggregation improves performance, we compute the final score by assigning different weights to each criterion based on its individual correlation with human judgment. As shown in Table~\ref{tab:weight}, the weighted configuration (0.6:1.2:1.2:1.0 for C1:C2:C3:C4) yields a slight improvement in average correlation compared to equal weighting (0.462 vs. 0.459). This can be viewed as an optional variant that offers marginal performance gains at the cost of increased metric complexity.

\begin{table}[t]
\begin{adjustbox}{max width=\linewidth}
\begin{tabular}{l|ccc|c}
\toprule
\textbf{Weight}  & $r_p$ & $r_s$ & $\tau$ & Average \\ \midrule
1 : 1 : 1 : 1                 & 0.497      & 0.489      & 0.390      & 0.459   \\
\textbf{0.6 : 1.2 : 1.2 : 1.0} & \textbf{0.498}    & \textbf{0.493}     & \textbf{0.394}         & \textbf{0.462}   \\ \bottomrule
\end{tabular}
\end{adjustbox}
\caption{Comparison of correlation scores using python benchmark between equal weighting and weighted aggregation for criteria (C1: C2: C3: C4).}
\label{tab:weight}
\end{table}

\subsection{Prompt sensitivity} \label{C.1}
In Section~\ref{5.5}, we experiment with the stability of the G-Eval and ReFEree methods under different prompt settings. We designed the following prompt variations for both the G-Eval and ReFEree prompts.
The system prompts for both G-Eval and ReFEree consist of three elements: instruction, evaluation criteria, and evaluation steps. We conduct experiments by constructing three combinations of these elements: instruction + evaluation criteria (ver1), instruction + evaluation steps (ver2), and instruction + evaluation criteria + evaluation steps (ver3).
Zero-shot LLMs are sensitive to the template used, meaning that changes in the tokens within the template can significantly impact performance. Table~\ref{tab:sensitivity} shows that our evaluation method achieves consistent performance even when the prompt is modified.

\begin{table}[t]
\begin{adjustbox}{max width=\linewidth}
\begin{tabular}{c|l|ccc|c}
\toprule
                        & \multicolumn{1}{c|}{Methods} & $r_p$        & $r_s$       & $\tau$        & Average            \\ \midrule
\multirow{3}{*}{Python} & ver1                        & 0.489          & 0.470          & 0.375          & 0.445          \\
                        & ver2                        & 0.492          & 0.480          & 0.385          & 0.452          \\
                        & \textbf{ver3*}        & \textbf{0.497}          & \textbf{0.489}          & \textbf{0.390}          & \textbf{0.459}          \\ \midrule
\multirow{3}{*}{Java}   & ver1                        & 0.504          & 0.488          & 0.414          & 0.469          \\
                        & ver2                        & 0.494          & 0.471          & 0.397          & 0.454          \\
                        & \textbf{ver3*}        & \textbf{0.515} & \textbf{0.502} & \textbf{0.423} & \textbf{0.480} \\ \bottomrule
\end{tabular}
\end{adjustbox}
\caption{Results of coefficient with varying instruction templates in ReFEree. \textbf{ver3*} is the final prompt used in our evaluation method.}
\label{tab:sensitivity}
\end{table}

\subsection{Stability Across Summary Lengths}
We additionally analyzed how ReFEree's evaluation results vary with summary length. To verify whether summary length affects correlation with human judgments, we divided the 1,825 Python benchmark samples into four groups based on summary length and measured the correlation between ReFEree scores and human labels within each group.

\begin{figure}[t]
    \centering
    \includegraphics[width=\linewidth]{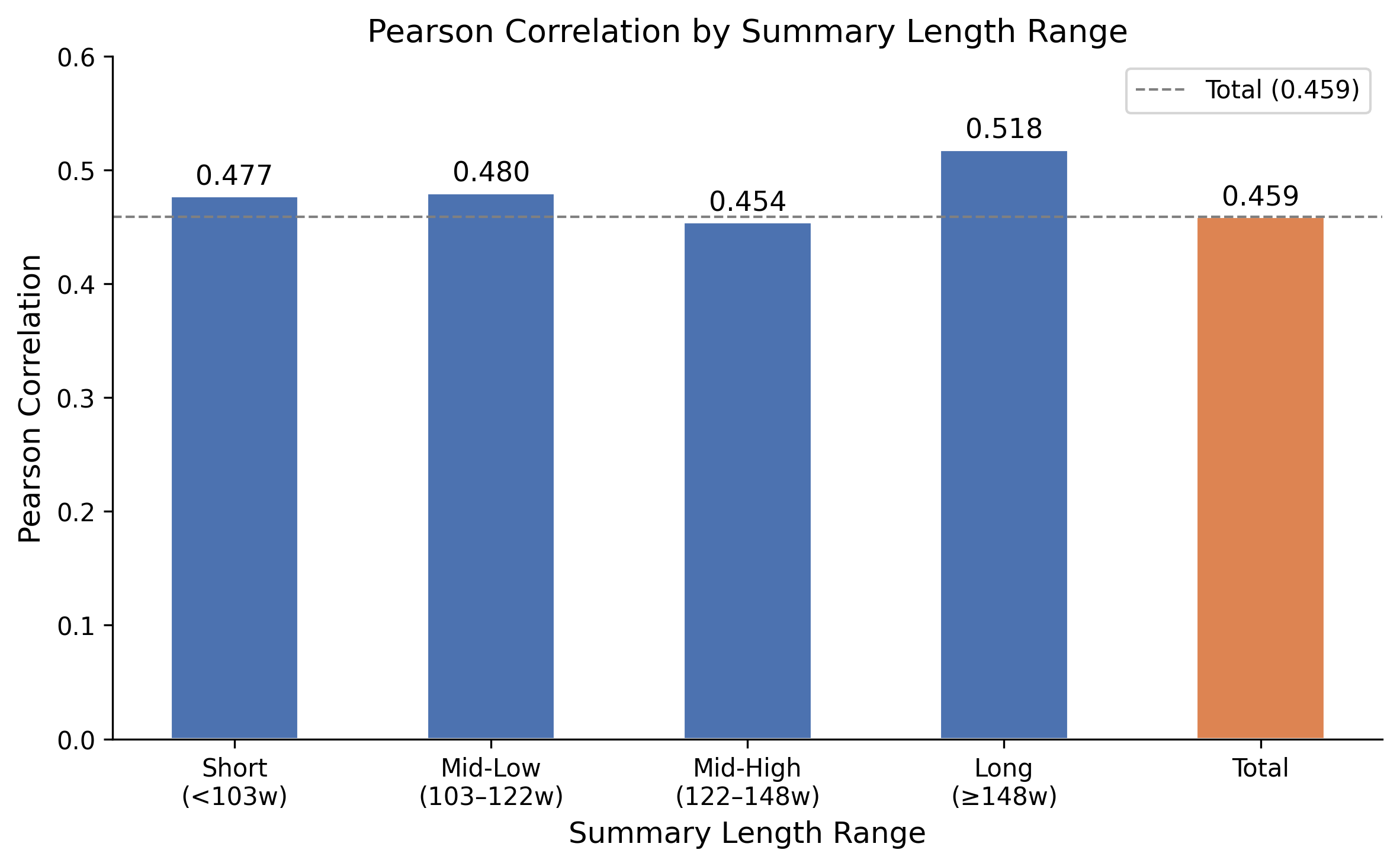}
    \caption{Stability of ReFEree across summary lengths}
    \label{figure:ablation_barchart1}
\end{figure}

The Figure~\ref{figure:ablation_barchart1} show that the overall correlation (0.459) remains consistently stable across all groups, ranging from approximately 0.45 to 0.52. This indicates that ReFEree maintains stable alignment with human judgment regardless of summary length. Since ReFEree measures the proportion of factually consistent content within a summary, the same error may result in different final scores depending on the summary length. However, this behavior is consistent with how humans evaluate summaries and does not introduce instability. The results show that this effect is not overly sensitive and remains well aligned with human judgments.

\subsection{Failure Analysis}
This section describes ReFEree's failure analysis.

\noindent \textbf{Error rate tends to increase as the number of dependency relations in the code grows. (Table~\ref{tab:ec1})} In particular, when the number of dependencies exceeds 11, the error rates for C3 and C4 rise noticeably. This trend suggests that as dependency information becomes more complex, making it harder to distinguish relevant signals from noise.

\begin{table}[h]
\begin{adjustbox}{max width=\linewidth}
\begin{tabular}{l|cccc|c}
\hline
\textbf{\#Dependencies} & \multicolumn{1}{l}{\textbf{C1}} & \multicolumn{1}{l}{\textbf{C2}} & \multicolumn{1}{l}{\textbf{C3}} & \multicolumn{1}{l|}{\textbf{C4}} & \multicolumn{1}{l}{\textbf{Average}} \\ \hline
0 & 3.8\% & 2.9\% & 2.6\% & 4.1\% & 3.4\% \\
1–2 & 3.1\% & 2.7\% & 3.8\% & 4.9\% & 3.6\% \\
3–5 & 5.2\% & 3.6\% & 4.3\% & 6.2\% & 4.8\% \\
6–10 & 5.7\% & 2.9\% & 5.1\% & 6.3\% & 5.0\% \\
11+ & 4.7\% & 2.4\% & 10.6\% & 9.4\% & 6.8\% \\ \hline
\end{tabular}
\end{adjustbox}
\caption{Error rate tends to increase as the number of dependency relations in the code grows.}
\label{tab:ec1}
\end{table}

\noindent \textbf{Error rates differ by dependency type. (Table~\ref{tab:ec2})} Internal dependencies (within the same file) show the lowest error rates, while cross-file and external API dependencies are higher. This is likely because internal dependencies rely on localized context, whereas cross-file and external cases require additional reasoning over dispersed information, increasing difficulty.

\begin{table}[h]
\begin{adjustbox}{max width=\linewidth}
\begin{tabular}{l|cccc|c}
\hline
\textbf{Dependency Type} & \textbf{C1} & \textbf{C2} & \textbf{C3} & \textbf{C4} & \textbf{Average} \\ \hline
Internal & 5.8\% & 3.4\% & 5.1\% & 5.8\% & \textbf{5.0\%} \\
Cross & 8.7\% & 2.4\% & 6.7\% & 9.1\% & \textbf{6.7\%} \\
External & 6.9\% & 4.9\% & 7.2\% & 8.0\% & \textbf{6.7\%} \\ \hline
\end{tabular}
\end{adjustbox}
\caption{Error Rates (\%) across different dependency type}
\label{tab:ec2}
\end{table}

\noindent \textbf{Error patterns across different dependency criteria. (Table~\ref{tab:ec3})} In particular, C1 (Name Inconsistency) and C4 (Context Irrelevance) show a higher tendency toward false positives (FP) compared to C2 and C3. For C4, the decision boundary is inherently more open to interpretation, which introduces a degree of subjectivity. In the case of C1, we observe that the model tends to apply a stricter interpretation of formal discrepancies (e.g., minor naming differences), leading to over-detection.

\begin{table}[h]
\begin{adjustbox}{max width=\linewidth}
\begin{tabular}{l|cccc}
\hline
\textbf{Metric} & \textbf{C1} & \textbf{C2} & \textbf{C3} & \textbf{C4} \\ \hline
Error Rate & 7.8\% & 3.8\% & 6.8\% & 7.6\% \\
FP (False Positive) & \textbf{6.5\%} & 0.9\% & 5.0\% & \textbf{6.4\%} \\
FN (False Negative) & 1.0\% & 2.9\% & 1.6\% & 1.2\% \\ \hline
\end{tabular}
\end{adjustbox}
\caption{Error patterns across different dependency criteria}
\label{tab:ec3}
\end{table}

\begin{table*}[h]
\centering
\begin{adjustbox}{max width=\textwidth}
\begin{tabular}{cccc}
\hline
\textbf{Input Code} & \textbf{Generated Summary} & \textbf{\begin{tabular}[c]{@{}c@{}}Summary \\ Label\end{tabular}} \\ \hline
\begin{minipage}[t]{0.7\linewidth}
\textbf{\#\# Python Input Code \#\#} \\
def iter\_slice (iterable: bytes, n: int): \\
\hspace*{2em}start = 0 \\
\hspace*{2em}stop = start + n \\
\hspace*{2em}final\_offset = len(iterable) \\
\hspace*{2em}while True: \\
\hspace*{4em}if start >= final\_offset:\\
\hspace*{6em}break\\
\hspace*{4em}rv = iterable[start:stop]\\
\hspace*{4em}start = stop\\
\hspace*{4em}stop = start + n\\
\hspace*{4em}yield rv, start >= final\_offset
\end{minipage} & 
\begin{minipage}[t]{0.6\linewidth}
The function iter\_slice is designed to process a bytes iterable by slicing it into fixed-size chunks of length n and yielding each chunk along with a boolean flag indicating whether the end of the iterable has been reached. Internally, it maintains a pair of indices, start and stop, which are used to extract the current slice from the iterable. 
After each yield, these indices are updated to prepare for the next slice.
\textcolor{red}{Notably, the function is named slice\_iterator, reflecting its role in iterating over slices of data \textbf{(C1, C3, C4)}.} 
\textcolor{red}{This approach is particularly useful in scenarios such as streaming data processing or chunked file reading, where handling large byte sequences in manageable pieces is essential \textbf{(C3, C4)}.} 
The boolean flag returned alongside each chunk helps client code determine when to stop requesting more data.
\end{minipage} & 
\begin{minipage}[t]{0.01\linewidth}
4
\end{minipage}  \\
\midrule
\begin{minipage}[t]{0.7\linewidth}
\textbf{\#\# Related Information \#\#} \\
\# itertools.chain \# \\
itertools.chain(*iterables): Make an iterator that returns elements from the first iterable until it is exhausted, then proceeds to the next iterable, until all of the iterables are exhausted. This combines multiple data sources into a single iterator. \\ \\
\textbf{\#\# Python Input Code \#\#} \\
def add\_ordereddicts(*odicts: OrderedDictType) -> OrderedDictType: \\
\hspace*{2em}items = [odict.items() for odict in odicts] \\
\hspace*{2em}return OrderedDictType(chain(*items))
\end{minipage} & 
\begin{minipage}[t]{0.6\linewidth}
The add\_ordereddicts function combines multiple OrderedDict instances into a single OrderedDict by merging their key-value pairs in sequence. 
It first extracts the items from each OrderedDict and then uses the itertools.chain method to create a flattened iterator over all these items. The function finally returns a new OrderedDictType constructed from this chained sequence, preserving the insertion order of keys from all input dictionaries. 
\textcolor{red}{This utility is particularly useful in scenarios where configuration settings or parameters from different sources need to be aggregated seamlessly \textbf{(C4)}.}
\textcolor{red}{Additionally, the function is designed to handle nested OrderedDicts by recursively flattening their contents before merging, ensuring a comprehensive consolidation of all nested key-value pairs \textbf{(C3, C4)}.}
\end{minipage} & 
\begin{minipage}[t]{0.01\linewidth}
4
\end{minipage}  \\ \midrule
\begin{minipage}[t]{0.7\linewidth}
\textbf{\#\# Java Input Code \#\#} \\
public static boolean isSameLength(final byte[] array1,final byte[] array2)\{ \\
\hspace*{2em}if (array1 == null \&\& array2 != null \&\& array2.length > 0 || array2 == null \&\& array1 != null \&\& array1.length > 0 || array1 != null \&\& array2 != null \&\& array1.length != array2.length) \{ \\
\hspace*{2em}return false; \\
  \} \\
  return true; \\
\}
\end{minipage} & 
\begin{minipage}[t]{0.6\linewidth}
\textcolor{red}{The Java function `checkEqualLength` takes two byte arrays as input and determines if they are of the same length. It returns a boolean value indicating whether the lengths of the two arrays are identical \textbf{(C1)}.} 
If either of the arrays is null while the other is non-null and has elements, or if both arrays are non-null but have different lengths, the function returns false. 
Otherwise, it returns true, indicating that the arrays are either both null or have the same length. 
\textcolor{red}{This utility is commonly used in data processing tasks where byte array comparisons are necessary, particularly in cryptographic applications and network data transmissions where array length uniformity is crucial \textbf{(C4)}.}
\end{minipage} & 
\begin{minipage}[t]{0.01\linewidth}
4
\end{minipage}  \\ \midrule
\begin{minipage}[t]{0.7\linewidth}
\textbf{\#\# Related Information \#\#} \\
\# java.lang.String \# \\
java.lang.String: The String class represents character strings in Java. All string literals are implemented as instances of this class. Strings are constant (immutable); their values cannot be changed after creation. The class provides methods for examining characters, comparing strings, searching, extracting substrings, and case conversions.
\\ \\
\# StringUtils.isEmpty \#\\
public static boolean isEmpty(final String str)\{\\
\hspace*{2em}return str == null || str.length() == 0;\\
\}\\
\\
\# StringUtils.INDEX\_NOT\_FOUND \#\\
public static final int INDEX\_NOT\_FOUND = -1;\\ \\
\textbf{\#\# Java Input Code \#\#} \\
public static int lastIndexOf(String str,String searchStr)\{ \\
\hspace*{2em}if (StringUtils.isEmpty(str)) \{ \\
\hspace*{4em}return StringUtils.INDEX\_NOT\_FOUND; \\
\hspace*{2em}\} \\
\hspace*{2em}return str.lastIndexOf(searchStr); \\
\}
\end{minipage} & 
\begin{minipage}[t]{0.6\linewidth}
\textcolor{red}{The `lastIndexOf` function is a utility method that finds the last occurrence of a given substring within a string. 
It utilizes the `StringUtils` class to first check if the input string is empty using the `isNullOrEmpty` method, which is actually not a real method in the `StringUtils` class \textbf{(C1, C2, C3, C4)}.} 
If the string is empty, it returns a constant `INDEX\_NOT\_FOUND` from the `StringUtils` class, which is correctly defined as `-1`. 
\textcolor{red}{This function is particularly useful in data processing and parsing scenarios where locating the position of a substring is critical \textbf{(C4)}.}
\textcolor{red}{Additionally, the method is often employed in web development for URL parsing and manipulation, though this context is not directly related to the core functionality of the code \textbf{(C3, C4)}.}
\end{minipage} & 
\begin{minipage}[t]{0.01\linewidth}
3
\end{minipage} \\ \bottomrule
\end{tabular}
\end{adjustbox}
\caption{\textbf{Sample from our benchmark.} Each summary is annotated with segment-level factual inconsistency labels (C1–C4) as defined in Section~\ref{3.1}, along with a summary-level label ranging from 1 to 5.}
\label{tab:benchmark_example}
\end{table*}

\begin{table*}[h]
\centering
\resizebox{\linewidth}{!}{
\begin{tabular}{ll}
\toprule
\textbf{Metric} & \textbf{Explanation}  \\ \midrule
Code & 
\begin{minipage}[t]{0.8\linewidth}
def convert\_to\_idn(url):\\
\hspace*{2em}parts = list(urllib.parse.urlsplit(url))\\
\hspace*{2em}try:\\
\hspace*{4em}parts{[}1{]}.encode(`ascii')\\
\hspace*{2em}except UnicodeEncodeError:\\
\hspace*{4em}host = parts{[}1{]}.rsplit(`:', 1)\\
\hspace*{4em}newhost = {[}{]}\\
\hspace*{4em}port = `'\\
\hspace*{4em}if len(host) == 2:\\
\hspace*{6em}port = host.pop()\\
\hspace*{4em}for h in host{[}0{]}.split(`.'):\\
\hspace*{6em}newhost.append(h.encode(`idna').decode(`utf-8'))\\
\hspace*{4em}parts{[}1{]} = `.'.join(newhost)\\
\hspace*{4em}if port:\\
\hspace*{6em}parts{[}1{]} += `:' + port\\
\hspace*{6em}return urllib.parse.urlunsplit(parts)\\
\hspace*{4em}else:\\
\hspace*{6em}return url \end{minipage} \\ \midrule
Generated Summary & 
\begin{minipage}[t]{0.8\linewidth}
The `convert\_to\_idn` function is designed to convert the hostname portion of a URL into its Internationalized Domain Name (IDN) ASCII-compatible encoding using the Punycode standard. It first splits the URL into its components using the `urlparse` function, then attempts to encode the hostname in ASCII. If the hostname contains Unicode characters, it separates the hostname from the port if present, encodes each label of the hostname into IDNA format, and finally recombines the parts, preserving the port number. The function returns a list of URL components with the converted hostname. This utility is particularly useful in web crawlers and multilingual domain management systems where URLs need to be normalized for consistent processing and storage.\end{minipage} \\ \midrule
Human Label & 0.6 \\ \midrule
ROUGE-L & 0.1831 \\ 
BERTScore & 0.8149 \\ 
SIDE & 0.7039 \\ 
\textbf{\textit{ReFEree}} & 
\begin{minipage}[t]{0.8\linewidth}
\textbf{Score: 0.55} \\
The `convert\_to\_idn` function is designed to convert the hostname portion of a URL into its Internationalized Domain Name (IDN) ASCII-compatible encoding using the Punycode standard. \textcolor{red}{It first splits the URL into its components using the `urlparse` function, then attempts to encode the hostname in ASCII \textbf{(C1, C2, C3, C4)}.} If the hostname contains Unicode characters, it separates the hostname from the port if present, encodes each label of the hostname into IDNA format, and finally recombines the parts, preserving the port number. \textcolor{red}{The function returns a list of URL components with the converted hostname \textbf{(C1, C2, C3, C4)}}. \textcolor{red}{This utility is particularly useful in web crawlers and multilingual domain management systems where URLs need to be normalized for consistent processing and storage \textbf{(C4)}.}\end{minipage} \\ \bottomrule
\end{tabular}}
\caption{\textbf{Additional examples of \textit{ReFEree} and existing methods:} Human labels for the code and generated summary are normalized to values between 0 and 1 to allow comparison with other metrics. Factual inconsistencies between the project-level code and the generated summary are highlighted in \textcolor{red}{red}. C1–C4 represent the individual factual inconsistency criteria described in Table~\ref{tab:cases_example}.}
\label{tab:cases_example_1}
\end{table*}

\begin{table*}[h]
\centering
\resizebox{\linewidth}{!}{
\begin{tabular}{ll}
\toprule
\textbf{Metric} & \textbf{Explanation}  \\ \midrule
Code & 
\begin{minipage}[t]{0.8\linewidth}
def system(cmd, data=None): \\
\hspace*{2em}import subprocess\\
\hspace*{2em}s = subprocess.Popen(cmd, shell=True, stdout=subprocess.PIPE, \\
\hspace*{4em}stdin=subprocess.PIPE)\\
\hspace*{2em}out, err = s.communicate(data)\\
\hspace*{2em}return out.decode('utf8')\end{minipage} \\ \midrule
Generated Summary & 
\begin{minipage}[t]{0.8\linewidth}
The `system` function serves as a lightweight wrapper around the `subprocess` module to execute shell commands. It launches a new process with the specified command string, optionally sending input data via the standard input stream. Internally, it uses the `execute\_command` function, which manages process creation and communication asynchronously, allowing for non-blocking interaction with the subprocess. The function captures both standard output and error streams, decodes the output using UTF-8 encoding, and returns it as a list of strings representing each line of the command's output. This design makes it particularly suitable for integration into machine learning workflows where real-time command execution and output parsing are required.\end{minipage} \\ \midrule
Human Label & 0.4 \\ \midrule
ROUGE-L & 0.1898 \\ 
BERTScore & 0.8297 \\ 
SIDE & 0.9999 \\ 
\textbf{ReFEree} & 
\begin{minipage}[t]{0.8\linewidth}
\textbf{Score: 0.6} \\
The `system` function serves as a lightweight wrapper around the `subprocess` module to execute shell commands. It launches a new process with the specified command string, optionally sending input data via the standard input stream. \textcolor{red}{Internally, it uses the `execute\_command` function, which manages process creation and communication asynchronously, allowing for non-blocking interaction with the subprocess \textbf{(C1, C3, C4)}.} \textcolor{red}{The function captures both standard output and error streams, decodes the output using UTF-8 encoding, and returns it as a list of strings representing each line of the command's output \textbf{(C2, C3, C4)}.} \textcolor{red}{This design makes it particularly suitable for integration into machine learning workflows where real-time command execution and output parsing are required \textbf{(C3, C4)}.}\end{minipage} \\ \bottomrule
\end{tabular}}
\caption{\textbf{Additional examples of \textit{ReFEree} and existing methods:} Human labels for the code and generated summary are normalized to values between 0 and 1 to allow comparison with other metrics. Factual inconsistencies between the project-level code and the generated summary are highlighted in \textcolor{red}{red}. C1–C4 represent the individual factual inconsistency criteria described in Table~\ref{tab:cases_example}.}
\label{tab:cases_example_2}
\end{table*}